\documentclass{article}

     \PassOptionsToPackage{numbers, compress}{natbib}
 \usepackage[preprint]{neurips_2026}

\newcommand{\Observation}[0]{\mathbf{X}}
\newcommand{\ActionQueries}[0]{\mathbf{{Q}_{action}}}
\newcommand{\RAMInput}[0]{\mathbf{\hat{Q}_{action}}}
\newcommand{\RAMInfo}[1]{\mathbf{Q}^{#1}_{\mathbf{RAM}}}
\newcommand{\InputActionQueries}[0]{\mathbf{{Q}}^{t}_{\mathbf{RAMaction}}}

\newcommand{\Prediction}[0]{\mathbf{O_{traj}}}
\newcommand{\PathWaypoints}[0]{\mathbf{O_{path}}}
\newcommand{\PathWaypointsi}[1]{\mathbf{O}^{#1}_{\mathbf{path}}}
\newcommand{\SpeedWaypoints}[0]{\mathbf{O}_{\mathbf{speed}}}
\newcommand{\SpeedWaypointsi}[1]{\mathbf{O}^{#1}_{\mathbf{speed}}}

\newcommand{\textinput}[0]{\mathbf{T}}

\newcommand{\textoutput}[0]{\mathbf{O_{text}}}
\newcommand{\textoutputi}[1]{\mathbf{O}^{#1}_{\mathbf{text}}}

\newcommand{\textoutputj}[0]{\mathbf{O}^{(j)}_\mathbf{text}}

\newcommand{\textinputj}[0]{\mathbf{T}^{(j)}}

\usepackage[acronym,nowarn,section,nogroupskip,nonumberlist]{glossaries}

\glsdisablehyper{}
\newacronym{RAM}{RAM}{Recurrent Action Memory}
\newacronym{DS}{DS}{driving score}
\newacronym{SR}{SR}{success rate}
\newacronym{AD}{AD}{autonomous driving}
\newacronym{SOTA}{SOTA}{state-of-the-art}
\newacronym{FM}{FM}{foundation model}
\newacronym{CoT}{CoT}{chain-of-thought}
\newacronym{LLM}{LLM}{large language model}
\newacronym{VLM}{VLM}{vision language model}
\newacronym{VLA}{VLA}{vision language action model}
\newacronym{BEV}{BEV}{bird's-eye view}
\newacronym{VQA}{VQA}{visual question answering}
\newacronym{E2E}{E2E}{end-to-end}
\newacronym{MLP}{MLP}{multi-layer perceptron}
\newacronym{AR}{AR}{autoregressive}
\newacronym{LoRA}{LoRA}{low-rank adaptation}
\newacronym{RAG}{RAG}{retrieval-augmented generation}
\newacronym{ADE}{ADE}{average displacement error}
\newacronym{OOD}{OOD}{out-of-distribution}
\usepackage[utf8]{inputenc} 
\usepackage[T1]{fontenc}    
\usepackage{hyperref}       
\usepackage{url}            
\usepackage{booktabs}       
\usepackage{amsfonts}       
\usepackage{nicefrac}       
\usepackage{microtype}      
\usepackage{xcolor}         
\usepackage{multirow}
\usepackage{wrapfig}
\usepackage{pifont}
\usepackage{graphicx}
\usepackage{subcaption}

\newcommand{\cmark}{\ding{51}}%
\newcommand{\xmark}{\ding{55}}%
\definecolor{forestgreen}{rgb}{0.13, 0.55, 0.13}
\newcommand{\imp}[1]{\textcolor{forestgreen}{\small{(+#1)}}}
\newcommand{\nimp}[1]{\textcolor{red}{\small{(-#1)}}}
\usepackage{algorithm}
\usepackage{algpseudocode}
\usepackage{cleveref} 

\crefname{section}{Sec.}{Secs.}
\crefname{figure}{Fig.}{Figs.}
\Crefname{section}{Section}{Sections}
\crefname{table}{Tab.}{Tabs.}
\Crefname{table}{Table}{Tables}

\newcommand{\blockcomment}[1]{}

\usepackage{colortbl}

\usepackage{makecell}

\definecolor{lightpurple}{RGB}{150,100,200}

\usepackage{etoolbox} 
\makeatletter
\newcommand\blfootnote[1]{%
  \begingroup
  \renewcommand\thefootnote{}%
  \@ifpackageloaded{hyperref}{\NoHyper}{}
    \footnote{#1}%
  \@ifpackageloaded{hyperref}{\endNoHyper}{}
  \addtocounter{footnote}{-1}
  \endgroup
}
\makeatother

\title{FIVE-VLA: Fast and EffectIVE Autonomous Driving with Recurrent Action Memory} 

\author{%
  Kemal Oksuz$^{1,2}$, Alexandru Buburuzan$^{1,2}$, Yuhan Yao$^{1}$, Puneet K. Dokania$^{1,2}$\thanks{The authors are from the ADAS Systems, Software \& Services Business Unit of Robert Bosch GmbH.
    } \\[0.5ex]
  $^{1}$Robert Bosch GmbH, Germany \quad
  $^{2}$Five AI Ltd., United Kingdom \\[0.3ex]
  \small\texttt{\{kemal.oksuz,alexandru.buburuzan,dokania.puneet\}@bosch.com} \\
  \small\texttt{yuhan.yao@de.bosch.com}
}

\begin{document}

\pagestyle{plain} 

\maketitle
\thispagestyle{plain} 

\begin{abstract}
State-of-the-art vision-language-action models (VLA) for autonomous driving face critical limitations: excessive parameter counts, inefficient high-resolution image processing, and lack of temporal memory. We introduce \textbf{Fast and EffectIVE VLA (FIVE-VLA)} to address these through two key contributions. First, we employ an efficient vision encoder that processes high-resolution ($448 \times 896$) images while generating only 98 tokens, over $5\times$ fewer than existing approaches, and bypass text generation entirely for single-pass trajectory prediction. Second, we propose \textbf{Recurrent Action Memory (RAM)}, a lightweight module that conditions action prediction on previous action tokens, providing temporal context critical for manoeuvres such as overtaking and emergency braking. With only 641M parameters, FIVE-VLA completes \textit{$\sim$10\% more routes without traffic rule infractions} than the previous state-of-the-art VLA on the challenging \textit{Bench2Drive} closed-loop driving benchmark. Non-reactive open-loop simulation on the large-scale real-world NVIDIA Physical AI AV dataset shows 10.2\% and 7.7\% lower collision-violation rates than SimLingo in single- and four-view settings, respectively. Additionally, FIVE-VLA runs at $\sim$30 fps on an A100 and $\sim$4 fps on a T4 GPU (proxy to an edge device), representing an \textit{8--30$\times$ speedup} over previous methods. 
\end{abstract}

\blockcomment{
\section{Introduction}
\label{sec:introduction}
\Glspl{VLA} hold great promise for \gls{AD}, offering rich semantic understanding and strong generalisation through large-scale pre-training. However, \gls{SOTA} \glspl{VLA}~\cite{ORION,autovla,feedbackguided,S4Driver,omnidrive,drivemlm,lmdrive,drivemoe,OpenDriveVLA,emma} adapted to driving contexts through domain-specific fine-tuning of \glspl{FM}~\cite{clip,qwen,internvl,chatgpt,gpt4,llava,gemini} introduce critical limitations regarding computational inefficiency and the absence of memory mechanisms. 

Despite the requirement for \glspl{VLA} to operate on vehicular edge devices, their efficiency remains suboptimal mainly due to three compounding factors. First, many \glspl{VLA} generate text during inference, either \gls{CoT} reasoning~\cite{autovla,feedbackguided,SimLingo} or superfluous tokens such as ``waypoints:''~\cite{SimLingo}, which, requires multiple forward passes due to \gls{AR} generation, substantially increasing latency. Second, most models have large parameter counts (e.g., 7B)~\cite{ORION,autovla,feedbackguided,S4Driver,omnidrive,drivemlm,lmdrive,drivevlm,solve,dima}: ORION~\cite{ORION}, for instance, achieves merely $\sim$1 fps on an A100 GPU. Third, higher-resolution images can improve driving performance~\cite{vad}, but many existing \glspl{VLA} use vision encoders pretrained at relatively low resolutions ($224 \times 224$ to $448 \times 448$)~\cite{S4Driver,drivemlm,drivevlm,DriveGPT4v2,alpamayor1}. To retain high-resolution detail, SimLingo~\cite{SimLingo} processes a $448 \times 896$ image as two $448 \times 448$ tiles. Each tile contributes 256 vision tokens, resulting in 512 vision tokens, thereby increasing its computational cost and inference latency.

Furthermore, the driving task substantially benefits from temporal context, as illustrated by the emergency-braking scene in \cref{fig:teaser}(a). However, \textit{the majority of \glspl{VLA} lack memory mechanisms}~\cite{feedbackguided,S4Driver,omnidrive,SimLingo,DriveGPT4v2}. A simple alternative is to supply past ego waypoints~\cite{autovla,emma,alpamayor1}, but this can encourage extrapolation of recorded expert motion rather than scene understanding, limiting robustness when the history instead reflects the model's own closed-loop driving~\cite{wen2020copycat,codevilla2019limitations,dehaan2019causal}. A few alternative approaches operate in the vision domain (\cref{fig:teaser}(b)): sliding window~\cite{autovla,emma,drivevlm} concatenate historical vision tokens with the current frame, generally \textit{exceeding} the token counts seen during \gls{VLM} pre-training, while Flex~\cite{flex} and Q-Former-based~\cite{blip2} approaches~\cite{ORION,solve} compress vision tokens via \textit{additional} transformer blocks, effectively replacing the pre-trained vision-language adapter of the \gls{VLM}~\cite{llava,fastvlm}. These strategies fundamentally differ from the design choices of the underlying \gls{VLM} backbone used to build the \gls{VLA}, potentially undermining the transfer learning from the \gls{VLM} by requiring significant amount of data for fine-tuning~\cite{alpamayor1,flex}. This naturally motivates an alternative memory design that is \textit{better aligned with the pre-trained \gls{VLM}'s architecture} while capturing the temporal context essential for safe driving. 

\begin{figure*}[t]
        \centering
        \includegraphics[width=0.98\textwidth]{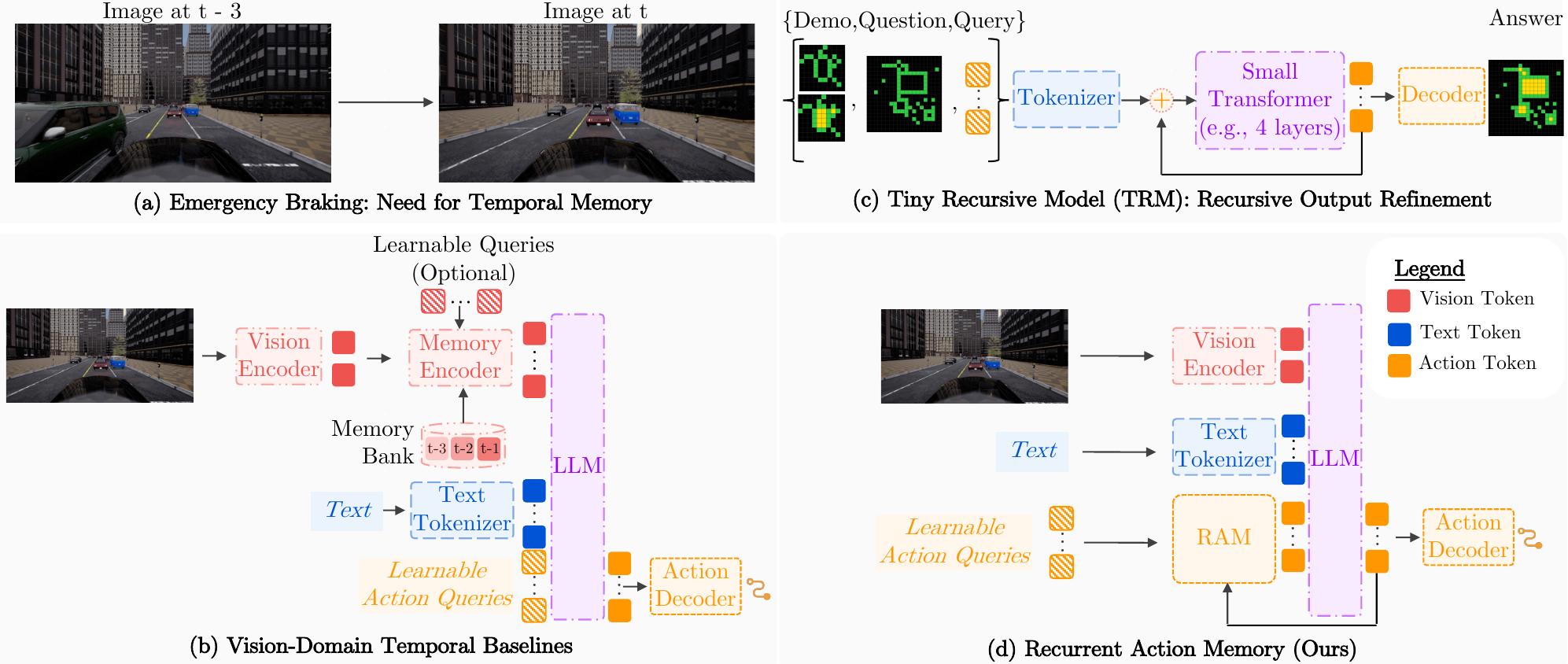}
        \vspace{-3pt}
        \caption{\textbf{(a) Emergency-braking scenario} illustrating the potential value of temporal context. A single frame may reveal a potential hazard, while successive observations can help assess the motion of other agents relative to the ego vehicle and the urgency of braking. \textbf{(b) Temporal vision baselines} (Flex, QT-Former, sliding window) fuse past and current frame features: Sliding Window concatenates frames without learnable queries, while Flex and QT-Former use self- and cross-attention, respectively. \textbf{(c) Tiny Recursive Model} iteratively refines predictions by conditioning on previously generated outputs, illustrated on ARC-AGI~\cite{arcprize2024technical} (few demonstrations followed by a question and a query for answer). \textbf{(d) Recurrent Action Memory (RAM)} extends this to \glspl{VLA} by conditioning action prediction on previous action tokens, operating in the \textit{action space} rather than the vision domain, preserving the integrity of \gls{VLM} (vision encoder, text tokenizer, \gls{LLM}).
        }
        \vspace{-7pt}
        \label{fig:teaser}
\end{figure*}

Recurrent modelling constitutes a particularly well-suited approach for designing memory mechanisms to process streaming video data~\cite{contextmattersrefiningobject,deeprecurrentconvolutionalnetworks,LRCN,recurrentvideomaskedautoencoders}, as is common in AD systems. Recent developments have demonstrated that recursive architectures, such as the Hierarchical Reasoning Model~\cite{hrm} and the Tiny Recursive Model (TRM)~\cite{trm}, can achieve exceptional performance on reasoning tasks including ARC-AGI~\cite{arcprize2024technical} and Sudoku puzzles. As shown in \cref{fig:teaser}(c), these models \textit{recursively refine their own solutions by conditioning on previously generated outputs and intermediate reasoning states}. The analogy to \gls{AD} is natural: just as these tasks require consistent reasoning over a fixed problem state, driving requires consistent decision-making over \textit{a slowly evolving scene}, where adjacent frames share strong similarity. However, the effectiveness of such approaches in streaming perception tasks within larger-scale \gls{VLA} frameworks remains unexplored.

In this paper, we introduce \textit{FIVE-VLA} (Fast and EffectIVE \gls{VLA} for \gls{AD}) to address the aforementioned limitations. FIVE-VLA uses FastVLM~\cite{fastvlm} as the backbone  \gls{VLM}, which strategically combines convolutional and attention operations to enable highly efficient processing of high-resolution images on edge devices, generating only 98 tokens from $448 \times 896$ images, a greater than $5\times$ compression over SimLingo~\cite{SimLingo}. Crucially, FIVE-VLA can bypass text generation entirely, directly predicting actions without \gls{CoT} reasoning or superfluous tokens, eliminating a key source of latency. Central to FIVE-VLA is \textit{\gls{RAM}}, a lightweight memory module that \textit{is aligned with the design choices of the \gls{VLM} backbone}, thereby maximising the benefits derived from the large-scale pre-training. As illustrated in \cref{fig:teaser}(d), \gls{RAM} conditions the current action prediction on the action tokens generated at the previous time step, establishing a recurrent feedback loop. Analogous to recursive modelling approaches~\cite{hrm,trm}, this mechanism \textit{reduces the model's reliance on inferring actions from scratch at each time step}, enabling more effective utilisation of the \gls{LLM} and yielding temporally consistent, higher-quality driving decisions. 

Our main contributions can be summarised as:

\begin{itemize}
    \item We propose \textbf{FIVE-VLA}, an efficient and effective \gls{VLA} achieving (i) approximately 77\% and 67\% Success Rate on the \textit{Bench2Drive} benchmark and \textit{Fail2Drive} generalisation split, respectively, \textit{outperforming SimLingo, the previous \gls{SOTA} \gls{VLA}, by an absolute margin of at least 10\% on both}; (ii) around 10\% less collision rate than SimLingo in open-loop non-reactive simulation on the large-scale \textit{real-world NVIDIA Physical AI AV} dataset; and (iii) $\sim$30 fps on A100 and 4 fps on T4, corresponding to $\sim$30$\times$ and $\sim$8$\times$ higher throughput compared to ORION and SimLingo, respectively.
    \item We introduce \textbf{\gls{RAM}}, a lightweight memory mechanism that recursively conditions action prediction on tokens from the previous time step. \gls{RAM} selectively improves capabilities where temporal reasoning is most critical, including \textit{giving way}, \textit{overtaking}, \textit{emergency braking}, and \textit{driving smoothness}.
\end{itemize}
\section{Background}

\textbf{Problem Formulation.} 
The observation set of \gls{AD} models typically comprises a subset of the following: (i) sensor readings $\Observation_{\text{sensor}}$, such as camera and LiDAR data; (ii) the ego vehicle state $\Observation_{\text{state}}$, including velocity and other dynamic parameters; and (iii) a route indicator $\Observation_{\text{route}}$, which can take the form of either GPS target coordinates or high-level commands such as \{\textit{go straight}, \textit{turn left}, \textit{turn right}\}. 
Each type of observation can encompass $T$ time steps, e.g., for the sensor readings, $\Observation_{\text{sensor}} = \{\Observation_{\text{sensor}}^i\}_{i=t-T+1}^t$.
We represent all observations by $\Observation = \{\Observation_{\text{sensor}}, \Observation_{\text{state}}, \Observation_{\text{route}}\}$.
Given $\Observation$, the \gls{AD} model is thus tasked with determining a safe trajectory $\Prediction \in \mathbb{R}^{k \times p}$ for the ego vehicle to follow at time step $t$, where $k$ represents the prediction horizon and $p$ denotes the dimensionality of the output at each time step (commonly $p=2$ for \gls{BEV}). 

\noindent\textbf{\gls{SOTA} in Autonomous Driving.} 
Two mainstream approaches currently dominate benchmark leaderboards~\cite{nuscenes,waymo,NAVSIM,bench2drive}. First, an \gls{E2E}-\gls{AD} model $\mathrm{f}_\theta(\cdot)$ directly produces the output trajectory $\Prediction$ from the observation set, i.e., $\Prediction=\mathrm{f}_\theta(\Observation)$,
where the sensory data typically includes either camera data alone~\cite{vad,uniad,vadv2,genad,Paradrive} or a combination of camera and LiDAR~\cite{gaussianfusion,DiffusionDrive,bridgedrive,transfuserplusplus}. Regardless of the sensing modality, these approaches generally encode sensory data in \gls{BEV} space, from which $\Prediction$ is obtained. 
Second, \glspl{VLA} employ \textit{\glspl{FM} as their backbone} and adapt them to driving tasks through fine-tuning. 
Different from \gls{E2E}-\gls{AD} models, \glspl{VLA} enhance driving performance by leveraging the learned representations of \glspl{FM} and exploit their language interface for various purposes, including \gls{CoT} reasoning and \gls{VQA}. 
We denote the input and output text sequences to the \gls{FM} as $\textinput$ and $\textoutput$, respectively, where each comprises a sequence of text tokens: $\textinput=\{\textinputj\}_{j=1}^n$ and $\textoutput=\{\textoutputj\}_{j=1}^m$, with $n$ and $m$ denoting their respective sequence lengths. While various architectural choices exist for constructing \glspl{VLA}~\cite{foundationmodelstrajectoryplanningreview}, a typical \gls{VLA} generates $\Prediction$ through a two-stage process. First, the text output $\textoutput$ is generated using the fine-tuned \gls{FM} $\mathrm{g}_\theta(\cdot,\cdot)$, as $\textoutput=\mathrm{g}_\theta( \Observation, \textinput)$, in an \gls{AR} manner, where each subsequent token of $\textoutput$ is conditioned on preceding tokens, requiring multiple passes through $\mathrm{g}_\theta(\cdot,\cdot)$. Second, $\Prediction$ is obtained by incorporating the text output: $\Prediction=\mathrm{h}_\phi(\mathrm{g}_\theta(\Observation, \textinput, \textoutput))$,
with $\mathrm{h}_\phi(\cdot)$ denoting the action decoder to yield $\Prediction$ from the representations of the \gls{FM}. These models typically rely exclusively on camera data and demonstrate strong performance. However, utilising a large model and the \gls{AR} generation of $\textoutput$ generally result in high inference latency.

\section{FIVE-VLA: Fast and EffectIVE VLA}
\begin{figure*}[t]
        \centering
        \includegraphics[width=0.98\textwidth]{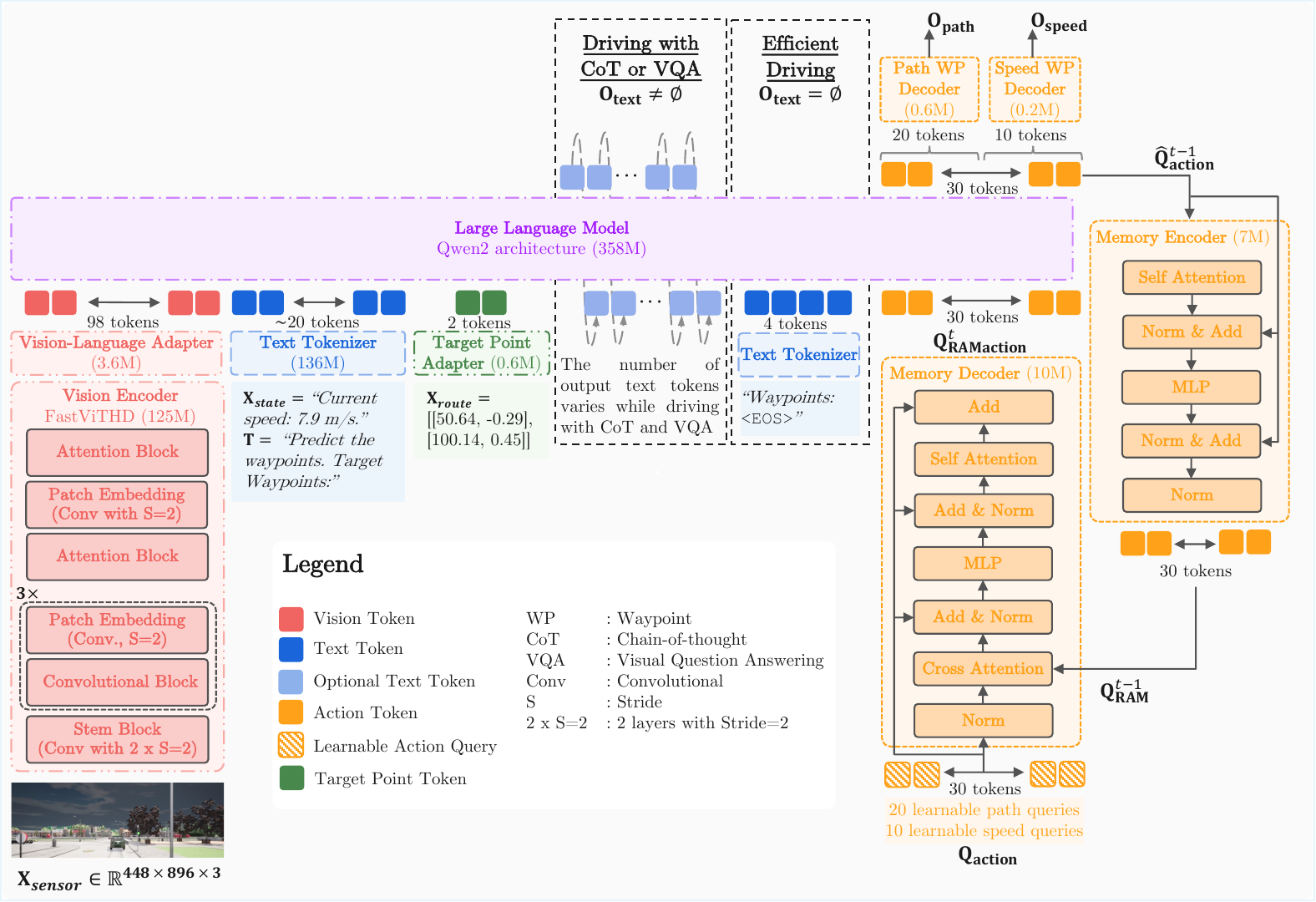}
        \vspace{-3pt}
        \caption{\textbf{Overview of FIVE-VLA, showing the single-view closed-loop configuration.} FIVE-VLA uses front image, ego velocity, target points and input text to produce path and speed waypoints. \gls{RAM} (orange blocks) conditions the action tokens $\ActionQueries$ on the output action tokens from the previous time step $\RAMInput^{t-1}$. It has 641M parameters, and yields the trajectory in a single pass using efficient driving mode.}
        \vspace{-5pt}
        \label{fig:overview}
\end{figure*}

To address the limitations of \glspl{VLA} in \gls{AD} regarding inefficiency and lack of temporal memory (\cref{sec:introduction}), FIVE-VLA incorporates three key design choices: (i) \textit{an efficient driving mode} predicting trajectories in a single forward pass without text generation (\cref{subsec:capabilities}), (ii) \textit{FastVLM~\cite{fastvlm} as the \gls{VLM} backbone} for efficient high-resolution image processing on edge devices (\cref{subsec:vlmbackbone}), and (iii) \textit{\gls{RAM}}, a novel lightweight memory module capturing temporal dependencies in the action domain (\cref{sec:recurrentactionmemory}).

\subsection{Overview of FIVE-VLA} 
\label{subsec:overview}
In the single-view configuration shown in \cref{fig:overview}, FIVE-VLA uses the front camera image $\Observation_{\text{sensor}} \in \mathbb{R}^{448 \times 896 \times 3}$, the ego velocity provided in text format $\Observation_{\text{state}}$, two subsequent target points for navigational conditioning $\Observation_{\text{route}} \in \mathbb{R}^{2 \times 2}$, and the text input $\textinput$ to produce $\textoutput=\mathrm{g}_\theta(\Observation, \textinput)$. $\mathrm{g}_\theta(\cdot, \cdot)$ includes vision encoder, vision-language adapter, text tokenizer, target point adapter and an \gls{LLM}.
Then, the learnable action queries $\ActionQueries$ are decoded as the trajectory through two components, $\Prediction=\{\PathWaypoints, \SpeedWaypoints\}$: (i) \textit{path waypoints} $\PathWaypoints \in \mathbb{R}^{N_{\text{path}} \times 2}$, representing positions in the ego frame sampled at 1-metre intervals along the path for steering control; and (ii) \textit{speed waypoints} $\SpeedWaypoints \in \mathbb{R}^{N_{\text{speed}} \times 2}$, representing ego-frame positions at fixed future time intervals. The waypoint counts and prediction horizons are dataset-specific. This can be expressed as:
\begin{align}
    \label{eq:FiveVLAtrajectory} 
\PathWaypoints, \SpeedWaypoints=\mathrm{h}_\phi(\mathrm{g}_\theta(\Observation, \textinput, \textoutput, \mathrm{RAM}_\psi(\ActionQueries, \RAMInput^{t-1}))),
\end{align}
where $\mathrm{RAM}_\psi(\cdot, \cdot)$ produces the action tokens fed to the \gls{LLM} using: (i) learnable action queries $\ActionQueries \in \mathbb{R}^{N \times d_{model}}$, where $d_{model}$ is the token embedding size and $N=N_{\text{path}}+N_{\text{speed}}$, and (ii) $\RAMInput^{t-1} \in \mathbb{R}^{N \times d_{model}}$ as the output action tokens of the \gls{LLM} from the previous time step $t-1$. At initialisation ($t=0$), since no prior \gls{RAM} information exists, we directly propagate  $\ActionQueries$ to the model, i.e., $\mathrm{RAM}_\psi(\ActionQueries, \RAMInput^{t-1}) = \ActionQueries$. This allows FIVE-VLA to operate both with and without \gls{RAM}, where \gls{RAM} serves to augment  $\ActionQueries$ with temporal context. Finally, $\mathrm{h}_\phi(\cdot)$ comprises path and speed waypoint decoders, each of which is an MLP processing corresponding representations in the predicted action tokens at current time step $\RAMInput^{t}$.

\subsection{Capabilities of FIVE-VLA}
\label{subsec:capabilities}
FIVE-VLA has three main capabilities: efficient driving without \gls{CoT}, driving with \gls{CoT} reasoning, and \gls{VQA} capabilities. To facilitate training of the \gls{RAM} across these capabilities, we structure training examples as temporal streams of length $T$ (comprising $T$ consecutive examples). Each stream, denoted as $\{\Observation^i, \textinput^i, \textoutputi{i}, \PathWaypointsi{i}, \SpeedWaypointsi{i}\}_{i=t-T+1}^t$, consists \textit{exclusively} of examples from a single capability, determined solely by the input text $\textinput$, as we detail below:

\textbf{Efficient Driving.} When $\textinput=$\textit{``Predict the waypoints.''}, SimLingo generates $\textoutput=$\textit{``Waypoints:''}, requiring four tokens to be produced by the \gls{LLM} in an \gls{AR} fashion (i.e., \texttt{``Way''}, \texttt{``points''}, \texttt{``:''}, \texttt{<EOS>}) prior to trajectory generation, increasing its inference latency. In contrast, recognizing that $\textoutput$ remains invariant across different inputs, we append these four tokens to the input token sequence. This simple insight bypasses generating $\textoutput$ entirely, enabling \textit{single-pass trajectory generation} via Eq.\eqref{eq:FiveVLAtrajectory} alone with $\textoutput=\emptyset$, and \textit{yielding the same trajectory output $(\PathWaypoints, \SpeedWaypoints)$}. Efficient driving is the default operational mode of FIVE-VLA due to its efficiency. 

\textbf{Driving with \gls{CoT} Reasoning.} If $\textinput=$\textit{``What should the ego do next?''}, the model first generates text $\textoutput$ containing \gls{CoT} reasoning, which subsequently informs trajectory generation in Eq.\eqref{eq:FiveVLAtrajectory}.

\textbf{\gls{VQA} Capability.} The model supports free-form queries through the text interface $\textinput$ to enhance model explainability. In this mode, FIVE-VLA generates text, $\textoutput$, prior to trajectory generation.

\subsection{VLM Backbone}
\label{subsec:vlmbackbone}
For \gls{AD} models to be deployable on vehicular platforms, they must efficiently process high-resolution images on edge devices. Existing \glspl{VLA} are commonly constrained by either their inability to leverage high-resolution images owing to the resolution constraints of their pretrained vision encoders, or their reliance on an excessive number of tokens, which impedes practical deployment. To address these limitations, we employ FastVLM~\cite{fastvlm} as our backbone, which is specifically designed to enable efficient processing of high-resolution images on edge devices while maintaining high accuracy. FastVLM combines FastViTHD vision encoder~\cite{fastvlm,fastvit} with Qwen2 \gls{LLM}~\cite{qwen2}, where the primary advantage of FastVLM lies in its vision encoder, which we elaborate upon next.

We utilise images with 448$\times$896 resolution. If processed by a ViT model~\cite{vit} as a vision encoder, which commonly uses 14$\times$14 patches~\cite{clip,vit,siglip}, such images yield 2048 tokens. Given this substantial token count, efficiency-oriented approaches aim to reduce the number of tokens to be passed to the \gls{LLM}~\cite{SimLingo,miniinternvl,prumerge,matryoshka}. For instance, Mini-Intern-VL~\cite{miniinternvl}, used as the backbone \gls{VLM} in SimLingo, employs \textit{pixel unshuffling}~\cite{pixelshuffle} within the InternViT-300M vision encoder to compress the token count to one-quarter of the original, though still yielding 512 tokens. Also, since the vision encoder is trained on 448$\times$448 images, processing a 448$\times$896 resolution image requires partitioning it into two separate tiles, thereby preventing cross-tile interactions within the vision encoder.

Alternatively, FastViTHD integrates convolutional~\cite{LeNet,MobileNet} and attention layers~\cite{attentionisallyouneed}, as illustrated in \cref{fig:overview}. In this architecture, a subset of the convolutional layers is employed with a stride of 2 to progressively reduce the spatial dimensions of the representations, thereby decreasing the number of tokens before the subsequent attention layers. This yields only 98 tokens from a 448$\times$896 image, representing approximately 21$\times$ and 5$\times$ reductions compared to the ViT and InternViT-300M, respectively. Also, FastViTHD comprises merely $125$M parameters, substantially fewer than the commonly employed ViT with 300M parameters (i.e., ViT-L). This architecture further enhances representation learning efficacy by eliminating the need for tile-based partitioning, thereby enabling global token interactions throughout the entire image within the vision encoder. 

To our knowledge, we are \textit{the first to tailor FastVLM} to a \gls{VLA} for \gls{AD}. For this, we use the smallest FastVLM with only around 620M parameters pretrained on images with resolution up to 1024$\times$1024, and fine-tune it by incorporating the \gls{AD}-specific modules (\cref{fig:overview}). The resulting FIVE-VLA  has only 641M parameters and runs more efficiently than its counterparts especially on edge devices.

\section{RAM: Recurrent Action Memory}
\label{sec:recurrentactionmemory}
The ability to leverage temporal information is essential for \gls{AD}, yet existing memory mechanisms operate in the vision domain, undermining the transfer learning from the pre-trained \gls{VLM}. We introduce \gls{RAM}, which instead operates in the \textit{action domain} by conditioning the current action prediction on the action tokens from previous time step, preserving the integrity of the \gls{VLM} (\cref{fig:teaser}(d)).

\noindent \textbf{\gls{RAM} Design.} The orange modules in \cref{fig:overview} illustrate the architecture of \gls{RAM}, which comprises two lightweight components. First, a memory encoder $\mathrm{RAMEncoder}(\cdot)$, one transformer block with self-attention, encodes the updated action queries from the previous time step $\RAMInput^{t-1}$ to obtain \gls{RAM} state $\RAMInfo{t-1}$,
\begin{align}
\label{eq:ramencoder}
\RAMInfo{t-1} = \mathrm{RAMEncoder}(\RAMInput^{t-1}).   
\end{align}
Second, a memory decoder $\mathrm{RAMDecoder}(\cdot,\cdot)$ conditions on $\RAMInfo{t-1}$ to yield the action tokens to be fed to the \gls{LLM} at time step $t$, $\InputActionQueries$. $\mathrm{RAMDecoder}(\cdot,\cdot)$ is another single transformer block with cross-attention, in which the action queries $\ActionQueries$ attend to the \gls{RAM} state $\RAMInfo{t-1}$,
\begin{align}
\label{eq:ramdecoder}
\InputActionQueries = \mathrm{RAMDecoder}(\ActionQueries, \RAMInfo{t-1}).
\end{align}
As shown in \cref{fig:overview} (orange blocks), since $\ActionQueries$ is not normalised by design, we apply layer normalisation~\cite{layernormalisation} prior to cross-attention to ensure magnitude compatibility with $\RAMInfo{t-1}$. Conversely, normalisation is not applied before feeding the updated action tokens into the \gls{LLM} to preserve their original characteristics. 
\gls{RAM} includes only 2 transformer blocks, which is notably smaller compared to the memory mechanisms in the vision domain, typically with 8 blocks~\cite{ORION,flex}.
Despite its simplicity in design, training a recursive module and using it during inference require careful design choices, which we detail in the following sections.

\begin{figure}[t]
    \centering
    \begin{subfigure}{0.45\textwidth}
        \centering
    \includegraphics[width=\textwidth]{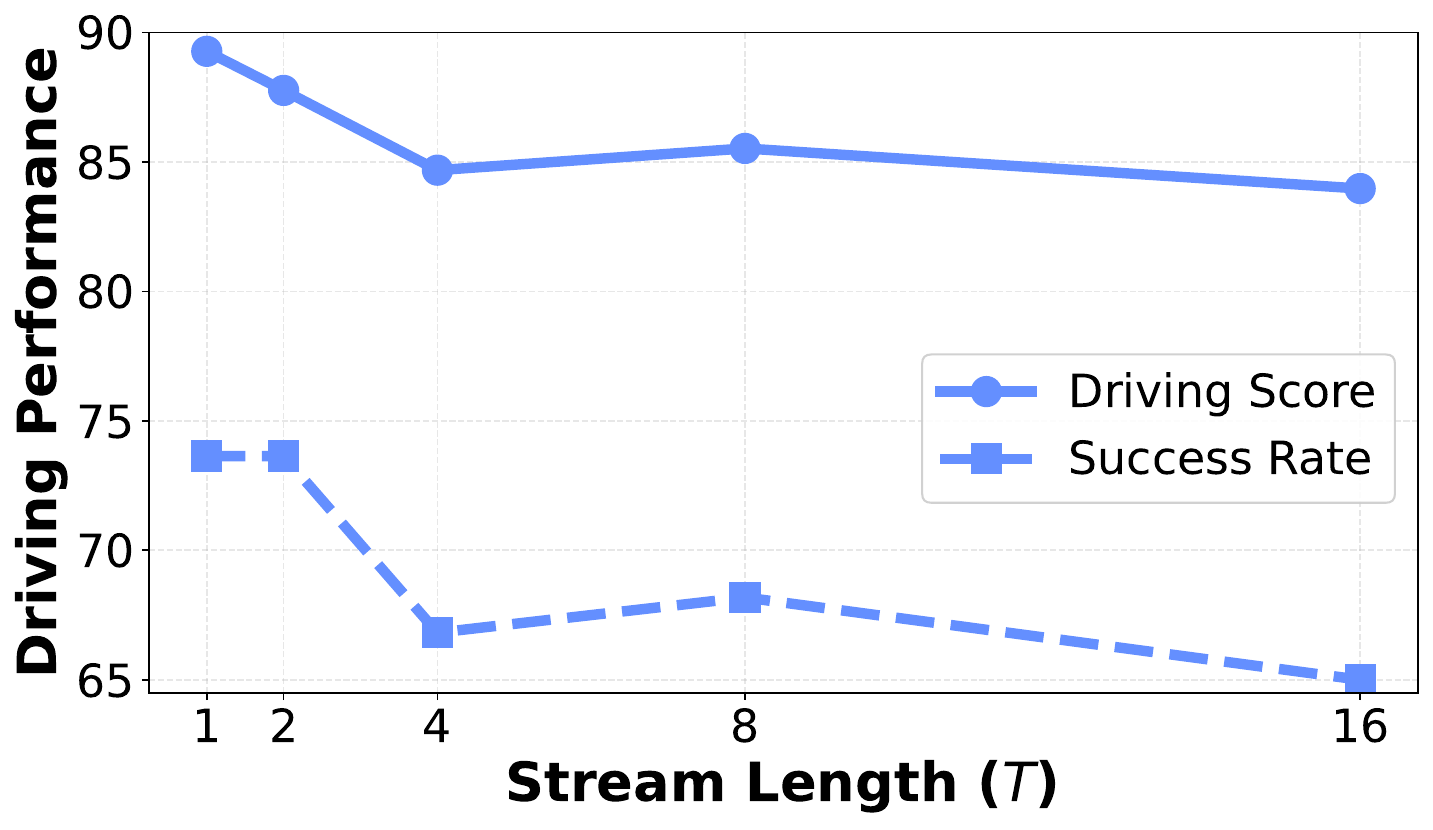}
        \caption{Changes in performance wrt. stream length.
        }
    \label{fig:performance_over_streamlength}        
    \end{subfigure}
    \hfill
    \begin{subfigure}{0.45\textwidth}
        \centering
        \includegraphics[width=\textwidth]{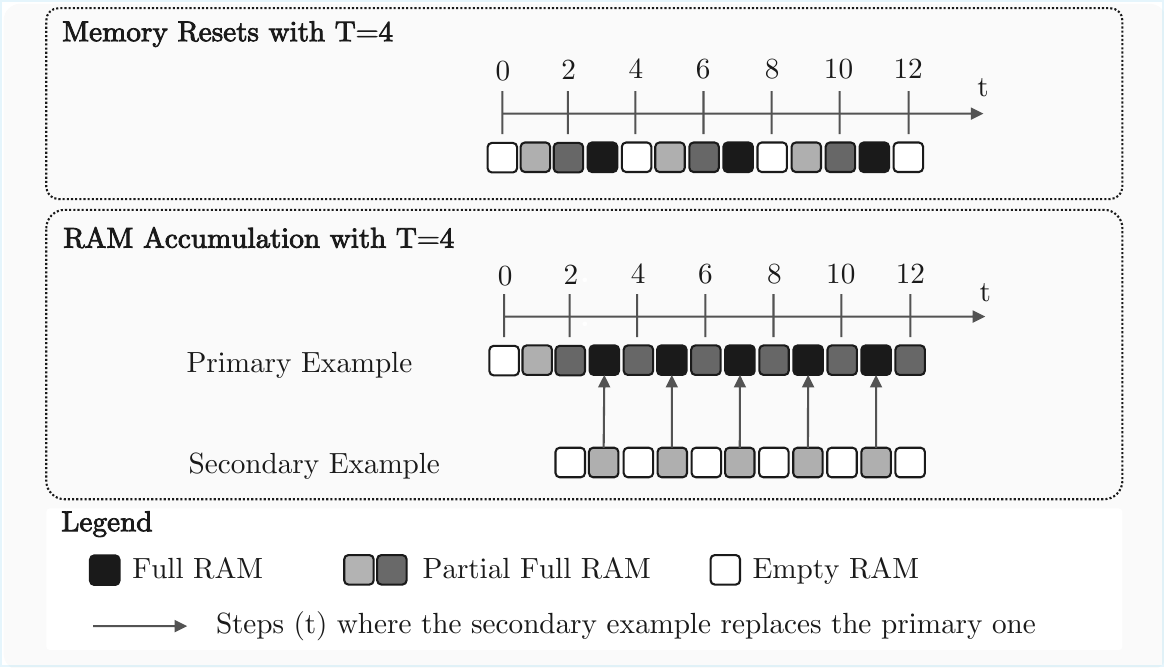}
        \caption{Different ways of using \gls{RAM} at inference}
        \label{fig:sub1}
    \end{subfigure}
    \vspace{-5pt}
    \caption{\textbf{(a)} Constructing batches from streaming data hurts the performance as stream length $T$ increases. \textbf{(b)} \gls{RAM} reset vs accumulation. \gls{RAM} accumulation prefills \gls{RAM} using a secondary example during inference before it replaces the primary one.
    }
    \label{fig:ramaccumulation}
    \vspace{-15pt}
\end{figure}

\subsection{Fine-tuning FIVE-VLA with \gls{RAM}}
\label{subsec:RAMfinetuning}
We now describe the training procedure for FIVE-VLA, which proceeds in two stages: first fine-tuning without \gls{RAM}, then incorporating \gls{RAM} in a second stage.

\textbf{Why fine-tune \gls{RAM}?}
Training models with memory mechanisms requires batches composed of sequential data streams, which inherently contain temporally-correlated observations. Such temporal correlation can degrade performance, primarily due to violations of the independent and identically distributed (IID) sampling assumption fundamental to stochastic gradient descent~\cite{mnih2015humanlevel,schaul2016prioritizedexperiencereplay,learningcontinuousvideostream,learningcontinuousvideostreamorthogonalgrad,iidbreak}. Correlated gradients reduce the informativeness of each training batch, as consecutive samples typically provide redundant information~\cite{learningcontinuousvideostreamorthogonalgrad}. This poses a significant challenge when training \glspl{VLA}, as GPU memory limitations often prevent increasing the batch size $B$ to offset reduced data diversity.

In our setting, a batch comprises $K$ distinct streams, each of length $T$, yielding $B=K \times T$ training examples. With 
$B$ fixed due to memory limitations, increasing the stream length $T$ necessitates a proportional decrease in the number of streams $K$, thereby reducing intra-batch variability and, consequently, gradient informativeness. This presents a trade-off between longer streams to enable memory mechanisms and  batch diversity.

To empirically quantify this effect, we train FIVE-VLA \textit{without \gls{RAM}} using data streams rather than IID sampling of individual observations, similar to Han et al.~\cite{learningcontinuousvideostreamorthogonalgrad}. We use SimLingo dataset~\cite{SimLingo} for training while systematically varying the stream length from 
$T=1$ (equivalent to IID sampling) to $T=16$, maintaining a constant batch size of $B=32$. We evaluate driving performance on the Bench2Drive closed-loop benchmark using both \gls{DS} and \gls{SR} metrics. As illustrated in \cref{fig:ramaccumulation}(a), both metrics degrade substantially as stream length increases beyond $T=2$.  
\textit{These results show that \glspl{VLA} trained with data streams could suffer from an inherent performance disadvantage attributable to reduced intra-batch variability. Therefore, to avoid potential performance degradation, we incorporate \gls{RAM} through fine-tuning after completing the initial training of FIVE-VLA without \gls{RAM}.}
  
\textbf{How to fine-tune \gls{RAM}?}  Proper initialisation of newly incorporated components is crucial for successful fine-tuning of \glspl{FM}~\cite{flamingo,lora}. For instance, parameter-efficient fine-tuning techniques~\cite{lora,qlora,adalora}, such as LoRA, typically learn a residual added to existing blocks in \glspl{FM}, expressed as $\mathbf{W} + \mathbf{\Delta W}$, where $\mathbf{W}$ represents the pretrained block parameters and $\mathbf{\Delta W}$ denotes the learned residual, commonly implemented as a matrix factorisation, e.g., $\mathbf{\Delta W}=\mathbf{B}\mathbf{A}$~\cite{lora}. To preserve pretrained knowledge and facilitate fine-tuning, $\mathbf{\Delta W}$ is typically initialised as $\mathbf{\Delta W} = \mathbf{0}$ at the start of the fine-tuning, e.g., by initialising $\mathbf{B}=\mathbf{0}$.

Similarly, to preserve the knowledge of FIVE-VLA after incorporating \gls{RAM} module, we ensure that $\InputActionQueries = \mathrm{RAM}_\psi(\ActionQueries, \RAMInput^{t-1}) = \ActionQueries$ at initialisation.
We achieve this through careful initialisation of the memory decoder (\cref{fig:overview}), specifically by setting the linear output projection layers of both cross-attention and self-attention mechanisms, as well as the second linear layer of the MLP, to zero. 
This ensures that, \textit{at the beginning of fine-tuning with \gls{RAM} modules, $\ActionQueries$ propagates exclusively through the skip connections}, while the action tokens generated in the last time step $\RAMInput^{t-1}$ do not contribute to $\InputActionQueries$. 
During \gls{RAM} fine-tuning, we optimise only the action decoders and \gls{LLM} along with the \gls{RAM} modules. While less critical than initialisation, we also employ gradient accumulation and stop gradient flow across time steps.

\subsection{Using \gls{RAM} at Inference: \gls{RAM} Accumulation}
It is well known that the sequence models, whether recursive or autoregressive, can exhibit poor generalisation beyond the sequence (or context) length on which they were trained~\cite{buitrago2025understanding,trainshorttestlong,contextwindowextension,streamingllm,StreamingVLM}. Due to the memory constraints of GPUs, the training sequence length is naturally set to a small number such as $T=4$. However, the driving scenarios are substantially longer than $T$, raising the question of how to utilise \gls{RAM} effectively at inference. 

One na\"ive approach is to \textit{reset \gls{RAM}} every $T$ time steps during inference, as shown in \cref{fig:ramaccumulation}(b). This ensures that the sequence length at inference is within the training sequence length $T$. However, \gls{RAM} becomes empty after each reset, thereby depriving the model of valuable temporal information. Alternatively, we introduce \textit{\gls{RAM} accumulation}, where we employ a secondary instance during inference to accumulate information in memory while the model relies on the primary instance to generate waypoints. At time $t$, the secondary instance is identical to the primary instance in terms of vision, language and target points, and only differs in RAM state. As shown in \cref{fig:ramaccumulation}(b), we initialise the secondary instance with an empty \gls{RAM}, and it becomes the primary one at every $T/2$ steps, where the primary example reaches the sequence length. 

\section{Experiments}
\label{sec:experiments}
Our experiments show that (i) FIVE-VLA outperforms existing \glspl{VLA} and \gls{E2E}-\gls{AD} methods; (ii) it is more efficient than its counterparts (up to $8\times$ faster than SimLingo); (iii) \gls{RAM} improves driving especially in \textit{safety-critical scenarios} requiring temporal understanding, such as emergency braking.

\textbf{Closed-Loop Experimental Setup.} We train FIVE-VLA on SimLingo dataset~\cite{SimLingo}, curated in CARLA~\cite{carla} with PDMLite expert~\cite{beisswenger2024pdmlite,drivelm} including $\sim$2M images at 4 fps, providing \textit{140 hours of driving data} with \gls{CoT} and \gls{VQA} annotations. For closed-loop driving, steering and acceleration are obtained via two PID controllers~\cite{pidcontroller} from path and speed waypoints. Unless otherwise noted, we use efficient driving with \gls{RAM} accumulation for inference, and report average results over three evaluation seeds. We evaluate on (i) \textit{Bench2Drive}~\cite{bench2drive}, a common benchmark, and (ii) \textit{Fail2Drive}~\cite{gerstenecker2026fail2drivebenchmarkingclosedloopdriving}, a recent  generalisation benchmark, both of which use (i) driving score (DS), route completion weighted by infraction penalties, and (ii) success rate (SR), proportion of routes completed without violations. 

\textbf{Open-Loop Experimental Setup.} To evaluate on real-world data, we train FIVE-VLA and SimLingo on the NVIDIA Physical AI AV dataset~\cite{Nvidiadataset}, using $\sim$860 hours for training and $\sim$350 hours for testing, with driving supervision only.  All models use only the driving data during training, i.e., no language supervision. During both training and evaluation, we perturb navigation target points derived from the recorded future ego trajectory with Student’s t distribution to mitigate expert-trajectory leakage while preserving coarse route guidance. We report average and final displacement errors (ADE/FDE) over a 3-second speed-waypoint horizon and their path counterparts over 30 metres, together with at-fault collision (NC-V) and time-to-collision violation (TTC-V) percentages from NAVSIM-style~\cite{NAVSIM} non-reactive simulation. Lower is better for all six metrics. Metric definitions are detailed in \cref{app:experiments}. 

\begin{table}[t]
\centering
\setlength{\tabcolsep}{0.4em}
\caption{\textbf{Comparison with \gls{SOTA}} on Bench2Drive and Fail2Drive generalisation benchmark. FIVE-VLA outperforms all approaches. M: Multi-view, S: Single view, L: LiDAR. Improvements (in \textcolor{forestgreen}{green} and \textcolor{red}{red}) are specified in comparison to the \underline{underlined} best camera-only baseline.}
\label{tab:sota}
\scalebox{0.59}{
\begin{tabular}{lcc|ccc|ccc|ccc|c}
\toprule
& & & \multicolumn{3}{c|}{\underline{\textbf{Bench2Drive}}} & \multicolumn{3}{c|}{\underline{\textbf{Fail2Drive In-Dist.}}} & \multicolumn{3}{c|}{\underline{\textbf{Fail2Drive Generalisation}}} & \\
\textbf{Method} & \textbf{Sensors} & \textbf{CoT} & \textbf{DS} $\uparrow$ & \textbf{SR(\%)} $\uparrow$ & \textbf{Mean Ability} $\uparrow$ & \textbf{DS} $\uparrow$ & \textbf{SR(\%)} $\uparrow$ & \textbf{HM} $\uparrow$ & \textbf{DS} $\uparrow$ & \textbf{SR(\%)} $\uparrow$ & \textbf{HM} $\uparrow$ & \textbf{Venue} \\
\hline
\multicolumn{13}{l}{\textbf{E2E-AD Approaches}} \\
UniAD~\cite{uniad} & M & \textemdash & 45.81 & 16.36 & 15.55  & 47.5 & 36.3 & 41.2 & 44.0 \small{(-7.4\%)} & 27.6 \small{(-24.0\%)} & 33.9 \small{(-17.6\%)}& CVPR23 \\
TCP~\cite{tcptraj} & S & \textemdash & 59.90 & 30.00 & 34.22 & 24.7 & 39.1 & 30.3 & 24.5 \small{(-0.8\%)} & 31.4 \small{(-19.7\%)} & 27.5 \small{(-9.1\%)} & NeurIPS22 \\
DriveTransformer~\cite{drivetransformer} & M & \textemdash & 63.46 & 35.01 & 38.60 &\textemdash&\textemdash&\textemdash&\textemdash&\textemdash&\textemdash& ICLR25 \\
Raw2Drive~\cite{raw2drive} & M & \textemdash & 71.36 & 50.24 & 53.34 & \textemdash&\textemdash&\textemdash&\textemdash&\textemdash&\textemdash& NeurIPS25 \\
VADv2~\cite{vadv2} & M & \textemdash & 76.15 & 50.46 & \textemdash &\textemdash&\textemdash&\textemdash&\textemdash&\textemdash&\textemdash & ICLR26 \\
PGS~\cite{PGS} & M & \textemdash & 78.08 & 48.64 & 53.40 &\textemdash&\textemdash&\textemdash&\textemdash&\textemdash&\textemdash & NeurIPS25 \\
GaussianFusion~\cite{gaussianfusion} & S, L & \textemdash & 79.10 & 54.40 & 56.30 &\textemdash&\textemdash&\textemdash&\textemdash&\textemdash&\textemdash& NeurIPS25 \\
Transfuser++~\cite{transfuserplusplus} & S, L & \textemdash & 84.21 & 67.27 & 64.39 &83.3 & 78.5 & 80.8 & 75.4 \small{(-9.5\%)} & 61.1 \small{(-22.2\%)} & 67.5 \small{(-16.5\%)}& ICCV23 \\
HiP-AD~\cite{hipad} & M & \textemdash &\underline{86.77}&\underline{69.09}&65.98& 74.1 & 70.7 & 72.4 & 67.1 \small{(-9.4\%)} & \underline{56.7} \small{(-19.8\%)} & 61.5 \small{(-15.1\%)} & ICCV25\\
BridgeDrive~\cite{bridgedrive} & S, L & \textemdash & 87.99 & 74.99 & 73.15 &\textemdash&\textemdash&\textemdash&\textemdash&\textemdash&\textemdash& ICLR26 \\
\hline
\multicolumn{13}{l}{\textbf{Vision-Language-Action Models}} \\
DriveMoE~\cite{drivemoe} & M & \xmark & 74.22 & 48.64 & 47.91 &\textemdash&\textemdash&\textemdash&\textemdash&\textemdash&\textemdash& CVPR26 \\
ORION~\cite{ORION} & M & \xmark & 77.74 & 54.62 & 54.72 & 53.0 & 52.0 & 52.5 & 51.2 \small{(-3.4\%)} & 46.0 \small{(-11.5\%)} & 48.5 \small{(-7.7\%)} & ICCV25 \\
AutoVLA~\cite{autovla} & M & \cmark & 78.84 & 57.73 & \textemdash &\textemdash&\textemdash&\textemdash&\textemdash&\textemdash&\textemdash& NeurIPS25 \\
SimLingo~\cite{SimLingo} & S & \cmark & 85.07 & 67.27 & \underline{67.03} & \underline{82.6} & \underline{79.3} & \underline{80.9} & \underline{71.7} \small{(-13.2\%)} & 55.0 \small{(-30.6\%)} & \underline{62.2} \small{(-23.1\%)} & CVPR25 \\
\hline
\rowcolor{blue!10}
FIVE-VLA & S & \xmark & \textbf{90.95} & \textbf{77.27} & \textbf{78.75} & \textbf{86.0} & \textbf{83.7} & \textbf{84.8} & \textbf{77.5} \small{(-9.9\%)} & \textbf{67.0} \small{(-20.0\%)} & \textbf{71.9} \small{(-15.3\%)}& Ours \\
\rowcolor{blue!10}
& & & \imp{4.18} & \imp{8.18} & \imp{11.72} & \imp{3.4} & \imp{4.4} & \imp{3.9} & \imp{5.8} & \imp{10.3} & \imp{9.7}& \\
\bottomrule
\end{tabular}
}
\vspace{-7pt}
\end{table}

\subsection{Comparison with State-of-the-art}

\textbf{Closed-Loop Driving on Bench2Drive.} \cref{tab:sota} compares FIVE-VLA with \gls{SOTA} on Bench2Drive benchmark with 220 challenging scenarios such as overtaking and merging~\cite{bench2drive}. Compared to SimLingo, the previous \gls{SOTA} \gls{VLA}, \textit{FIVE-VLA achieves a 10\% absolute gain in \gls{SR}} (77.27\% vs. 67.27\%), $\sim$6 points higher \gls{DS} (90.95 vs. 85.07), and a $\sim$12 points gain in mean ability score (see \cref{tab:sota_app} for individual abilities). These significant gains stem from two complementary factors: \gls{RAM} and using FastVLM, mainly helping for effective processing of high-resolution images without tile-based partitioning, unlike SimLingo (see \cref{subsec:ablation} for ablations). \textit{FIVE-VLA establishes a new \gls{SOTA}}, surpassing all \gls{VLA} and \gls{E2E}-\gls{AD} approaches. Despite using only a single front-facing camera, FIVE-VLA outperforms BridgeDrive, the leading camera-LiDAR approach, by $\sim$3 \gls{DS} and $\sim$2 \gls{SR}.

\textbf{Generalisation Performance on Fail2Drive.} Fail2Drive~\cite{gerstenecker2026fail2drivebenchmarkingclosedloopdriving}  stress-tests \gls{AD} models on 100 in-distribution (ID) and 100 generalisation scenarios  featuring \textit{\gls{OOD} hazards such as animals and unconventional vehicle parking},  reporting the harmonic mean (HM) of \gls{DS} and \gls{SR} as the final performance measure. As shown in Tab.~\ref{tab:sota}, \textit{FIVE-VLA establishes a new \gls{SOTA} across both splits}, outperforming SimLingo by 3.9 HM on the ID split. More critically, this gap widens to 9.7 HM on the generalisation split, indicating that the gains from RAM and effective high-resolution perception reflect genuinely stronger scene understanding rather than overfitting to the training distribution (refer to \cref{fig:fail2drive_examples} for qualitative examples). FIVE-VLA's smaller performance drop when transitioning from ID to generalisation compared to SimLingo (-9.9\% vs. -13.2\% in \gls{DS}; -20.0\% vs. -30.6\% in \gls{SR}) further supports this conclusion. Taken together with the Bench2Drive results, these consistent gains suggest that FIVE-VLA's closed-loop driving improvements are systematic rather than benchmark-specific. 

\begin{table*}[t]
\centering
\begin{minipage}[t]{0.55\textwidth}
\vspace{0pt}
\centering
\setlength{\tabcolsep}{0.15em}
\caption{\textbf{Single- and four-view results on NVIDIA Physical AI AV.} Errors in metres; NC-V/TTC-V: no at-fault collision and time to collision violation percentages on non-reactive NAVSIM-style simulation~\cite{NAVSIM}. \textbf{Bold}: best per view setting.}
\vspace{-5pt}
\label{tab:nvidiaav}
\resizebox{\linewidth}{!}{%
\begin{tabular}{@{}lcc|cc|cc|cc@{}}
\toprule
& &\textbf{Vision}& \multicolumn{2}{c|}{\textbf{Speed (3\,s)}} & \multicolumn{2}{c|}{\textbf{Path (30\,m)}} & \multicolumn{2}{c}{\textbf{Safety (\%)}} \\
\cmidrule(lr){4-5}\cmidrule(lr){6-7}\cmidrule(lr){8-9}
\textbf{Method} & \textbf{Views} & \textbf{Tokens} & \textbf{ADE} $\downarrow$ & \textbf{FDE} $\downarrow$ & \textbf{ADE} $\downarrow$ & \textbf{FDE} $\downarrow$ & \textbf{NC-V} $\downarrow$ & \textbf{TTC-V} $\downarrow$ \\
\midrule
SimLingo~\cite{SimLingo} & 1 & 512 & 0.550 & 1.466 & 0.218 & 0.420 & 0.87 & 1.26 \\
\makecell[l]{FIVE-VLA (w/o \gls{RAM})} & 1 & \textbf{98} & 0.549 & 1.464 & 0.213 & 0.405 & 0.84 & 1.25 \\
\rowcolor{blue!10}
FIVE-VLA & 1 & \textbf{98} & \textbf{0.402} & \textbf{1.128} & \textbf{0.207} & \textbf{0.392} & \textbf{0.78} & \textbf{1.16} \\
\midrule
SimLingo~\cite{SimLingo} & 4 & 2048 & 0.504 & 1.358 & 0.206 & 0.407 & 0.82 & 1.21 \\
\rowcolor{blue!10}
FIVE-VLA & 4 & \textbf{392} & \textbf{0.391} & \textbf{1.100} & \textbf{0.197} & \textbf{0.379} & \textbf{0.75} & \textbf{1.15} \\
\bottomrule
\end{tabular}}
\end{minipage}\hfill%
\begin{minipage}[t]{0.42\textwidth}
\vspace{0pt}
\centering
\setlength{\tabcolsep}{0.15em}
\caption{\textbf{Efficiency.} 
SimLingo is reported without \gls{CoT} for fair fps comparison, so the numbers slightly differ from \cref{tab:sota}. OOM: Out-of-memory. AutoVLA reports $1$ fps, the GPU type is not specified. }
\vspace{-5pt}
\label{tab:efficiency}
\resizebox{\linewidth}{!}{%
\begin{tabular}{@{}l|c|ccc|cc@{}}
\toprule
& & \multicolumn{3}{c|}{\makecell{\textbf{Throughput} \textbf{(fps $\uparrow$})}} & \multicolumn{2}{c}{\textbf{Accuracy}} \\
\textbf{Method} & \makecell{\textbf{\#Params}\\$\downarrow$} & \textbf{T4} & \textbf{A100} & \textbf{H200} & \makecell{\textbf{DS}\\$\uparrow$} & \makecell{\textbf{SR}\\\textbf{(\%)} $\uparrow$} \\
\hline
ORION~\cite{ORION} & 7B & OOM & 0.98 & 2.13 & 77.74 & 54.62 \\
AutoVLA~\cite{autovla} & 3B & \textemdash & $\sim$1 & \textemdash & 78.84 & 57.73 \\
SimLingo~\cite{SimLingo} & 1B & 0.50 & 7.99 & 14.14 & 85.88 & 68.18 \\
\hline
\rowcolor{blue!10}
FIVE-VLA & \textbf{641M} & \textbf{3.89} & \textbf{29.96} & \textbf{56.71} & \textbf{90.95} & \textbf{77.27} \\
\bottomrule
\end{tabular}}
\end{minipage}
\vspace{-12pt}
\end{table*}

\begin{figure}[t]
\centering
\includegraphics[width=\textwidth]{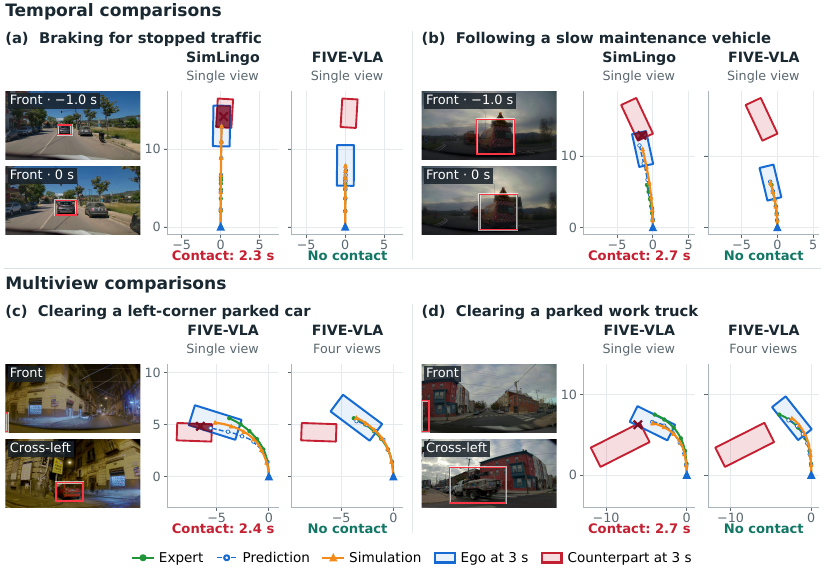}
\vspace{-7pt}
\caption{\textbf{Qualitative comparisons on NVIDIA Physical AI AV.} \textbf{Top:} (a) braking for stopped traffic and (b) following a slow maintenance vehicle, comparing single-view SimLingo and FIVE-VLA. \textbf{Bottom:} (c) clearing a parked car and (d) a work truck, comparing single- and four-view FIVE-VLA. Each panel shows two images and paired BEVs: temporal images are 1\,s apart (fixed crop in (a)); multiview images are front/cross-left. Footprints are synchronised at $+3$\,s; crosses mark overlap and red labels give first at-fault contact times. Rollouts are open-loop and non-reactive.}
\label{fig:nvidia_qualitative}
\vspace{-7pt}
\end{figure}

\textbf{Open-loop and Multi-View Evaluation on NVIDIA Physical AI AV.} To show that FIVE-VLA has competitive driving performance in the real world, we scaled FIVE-VLA to NVIDIA Physical AI AV dataset as the largest publicly available real-world driving dataset, capturing a diverse distribution of weather conditions, lighting, and complex interactions. \cref{tab:nvidiaav} shows that, without \gls{RAM}, FIVE-VLA performs slightly better than SimLingo (speed ADE: 0.549 vs.\ 0.550\,m; NC-V: 0.84\% vs.\ 0.87\%) using 98 rather than 512 vision tokens. As expected, adding \gls{RAM} mainly improves speed behaviour of the model (speed ADE gain by 26.7\%) with a milder benefit to path ADE (2.8\%). More crucially, safety of the model improves significantly: $7\%$ and $\sim10\%$ relative NC-V gain in comparison to FIVE-VLA (w/o RAM) and SimLingo. Four-view FIVE-VLA further reduces speed ADE (from 0.402 to 0.391\,m), path ADE (from 0.207 to 0.197\,m) and violation percentages. Compared with four-view SimLingo, it achieves 22.4\% lower speed ADE and 7.7\% lower NC-V with 392 rather than 2048 vision tokens. The qualitative examples in \cref{fig:nvidia_qualitative} illustrate differences in braking, following, and turning clearance, consistent with the aggregate results. These results complement our closed-loop findings and show that FIVE-VLA benefits from additional camera views while retaining its relative visual-token reduction.

\textbf{Efficiency.} We measure throughput as frames-per-second (fps) of the single-view FIVE-VLA and SimLingo models trained on the SimLingo dataset~\cite{SimLingo} on three GPU types: Tesla T4 (16GB VRAM, 130 INT8 TOPS) as \textit{a proxy for vehicular edge devices} (e.g., Qualcomm QC8775); A100 (80GB VRAM, 624-1248 INT8 TOPS), a mid-tier GPU; and H200 (141GB VRAM), a high-end GPU. \cref{tab:efficiency} shows that FIVE-VLA is consistently more efficient thanks to its \textit{efficient driving mode and FastVLM backbone}. On a Tesla T4, FIVE-VLA achieves 3.89 fps, \textit{$\sim$$8\times$ faster than SimLingo}, while ORION fails entirely due to out-of-memory. On higher-end GPUs, FIVE-VLA is nearly $4\times$ and $30\times$ faster than SimLingo and ORION on A100, and exceeds 50 fps on H200. With 641M parameters, FIVE-VLA is around 9\%, 21\% and 64\% the size of ORION, AutoVLA and SimLingo, respectively.

\subsection{Is Recurrent Action Memory Useful?}

\textbf{Contribution of \gls{RAM}.} To demonstrate the contribution of \gls{RAM}, we compare FIVE-VLA (including \gls{RAM}) with the FIVE-VLA variant obtained after the first stage training (without \gls{RAM}) in terms of overall performance and individual driving abilities. \cref{tab:rameffectonability} shows that \gls{RAM} yields notable gains in overall performance with +2.46 \gls{DS} and +4.24 \gls{SR}. Examining individual abilities reveals an expected pattern: \textit{capabilities that inherently benefit from temporal context show the most significant improvements.} Specifically, give-way scenarios exhibit the largest gain (+26.67 points), followed by overtaking (+7.40 points) and emergency braking (+6.11 points). These results align with the intuition that \gls{RAM}'s temporal modelling is particularly valuable for manoeuvres requiring anticipation over time. In contrast, merging and traffic sign recognition show minimal change (+0.83 and +1.22 points, respectively). This modest impact is anticipated, as merging performance is primarily limited by the lack of multi-view spatial information rather than temporal context, while traffic sign recognition can be effectively performed from static images. Last but not least, being informed by the previous action tokens, \gls{RAM} improves the smoothness metric substantially as well (more than 10\% relative gain). \textit{These findings validate that \gls{RAM} selectively enhances capabilities where temporal reasoning provides the most value, while having minimal effect on other abilities.}

\begin{table}[t]
\centering
\setlength{\tabcolsep}{0.5em}
\caption{\textbf{The contribution of \gls{RAM}.} \gls{RAM} notably improves abilities requiring temporal context (${>}5$\% relative gain), such as emergency brake, giving way and overtaking, and the smoothness (a.k.a. comfortness). Relative gain: $a/b-1$ where $a$ and $b$ are the performances with and without \gls{RAM}.}
\label{tab:rameffectonability}
\scalebox{0.65}{
\begin{tabular}{c|cc|cc|ccccc|c}
\toprule
& & & & & \multicolumn{6}{c}{\textbf{Ability}$\uparrow$} \\
\cline{6-11}
\textbf{\gls{RAM}} & \textbf{DS} $\uparrow$ & \textbf{SR(\%)} $\uparrow$ &\textbf{Efficiency} $\uparrow$&\textbf{Smoothness} $\uparrow$& \textbf{Merging} & \makecell{\textbf{Over-}\\\textbf{taking}} & \makecell{\textbf{Emergency}\\\textbf{Brake}} & \makecell{\textbf{Give}\\\textbf{Way}} & \makecell{\textbf{Traffic}\\\textbf{Sign}} & \textbf{Mean} \\ \hline
\xmark &88.49&73.03&\textbf{262.56}&25.90&59.17&76.30&81.67&50.00&84.39&70.30\\
\hline
\rowcolor{blue!10}
\cmark&\textbf{90.95}&\textbf{77.27}&255.07&\textbf{28.54}&\textbf{60.00}&\textbf{83.70}&\textbf{87.78}&\textbf{76.67}&\textbf{85.61}&\textbf{78.75} \\
\rowcolor{blue!10}(Absolute gain)&\imp{2.46}&\imp{4.24}&\nimp{7.49}&\imp{2.64}&\imp{0.83}&\imp{7.40}&\imp{6.11}&\imp{26.67}&\imp{1.22}&\imp{8.45} \\
\rowcolor{blue!10}(Relative gain)&\imp{2.78\%}&\imp{5.81\%}&\nimp{2.85\%}&\imp{10.19\%}&\imp{1.40\%}&\imp{9.70\%}&\imp{7.48\%}&\imp{53.34\%}&\imp{1.45\%}&\imp{12.02\%} \\
\bottomrule
\end{tabular}
}
\vspace{-7pt}
\end{table}

\begin{table*}[t]
\centering
\begin{minipage}[t]{0.51\textwidth}
\vspace{0pt}
\centering
\setlength{\tabcolsep}{0.15em}
\caption{\textbf{Explicit action-history baselines.} Single-view FIVE-VLA. History: recorded expert motion (open-loop) or model-controlled motion (closed-loop). Speed ADE/FDE: metres over 3\,s.}
\vspace{-4pt}
\label{tab:actionhistory}
\resizebox{\linewidth}{!}{%
\begin{tabular}{@{}cl|cc|cc|cc|cc@{}}
\toprule
& & \multicolumn{6}{c|}{\textbf{Closed-loop}} & \multicolumn{2}{c}{\textbf{Open-loop}} \\
& & \multicolumn{2}{c|}{\textbf{Bench2Drive}} & \multicolumn{2}{c|}{\textbf{Fail2Drive ID}} & \multicolumn{2}{c|}{\textbf{Fail2Drive Gen.}} & \multicolumn{2}{c}{\textbf{NVIDIA}} \\
\cmidrule(lr){3-4}\cmidrule(lr){5-6}\cmidrule(lr){7-8}\cmidrule(lr){9-10}
\textbf{\gls{RAM}} & \textbf{History input} & \textbf{DS} $\uparrow$ & \textbf{SR(\%)} $\uparrow$ & \textbf{DS} $\uparrow$ & \textbf{SR(\%)} $\uparrow$ & \textbf{DS} $\uparrow$ & \textbf{SR(\%)} $\uparrow$ & \textbf{ADE} $\downarrow$ & \textbf{FDE} $\downarrow$ \\
\midrule
\xmark & None & 88.49 & 73.03 & 83.5 & 78.0 & 75.9 & 61.0 & 0.549 & 1.464 \\
\xmark & 3 waypoints & 84.54 & 64.39 & 77.9 & 69.0 & 68.7 & 56.0 & 0.373 & 1.078 \\
\xmark & 10 waypoints & 85.95 & 70.00 & 76.1 & 67.0 & 73.2 & 60.0 & \textbf{0.358} & \textbf{1.038} \\
\midrule
\rowcolor{blue!10}
\cmark & None & \textbf{90.95} & \textbf{77.27} & \textbf{86.0} & \textbf{83.7} & \textbf{77.5} & \textbf{67.0} & 0.402 & 1.128 \\
\bottomrule
\end{tabular}}
\end{minipage}\hfill%
\begin{minipage}[t]{0.45\textwidth}
\vspace{0pt}
\centering
\setlength{\tabcolsep}{0.15em}
\caption{\textbf{Comparison with vision-domain temporal baselines.} \gls{RAM} has the best performance and is the smallest learned module. All methods use single-view FIVE-VLA. 
}
\vspace{-4pt}
\label{tab:othermemorytypes}
\resizebox{\linewidth}{!}{%
\begin{tabular}{@{}l|c|ccc|cc@{}}
\toprule
\multirow{2}{*}{\textbf{Approach}}
& \multirow{2}{*}{\textbf{\#Params} $\downarrow$}
& \multicolumn{3}{c|}{\textbf{Throughput (fps)} $\uparrow$}
& \multirow{2}{*}{\textbf{DS} $\uparrow$}
& \multirow{2}{*}{\textbf{SR(\%)} $\uparrow$} \\
&
& \textbf{T4}
& \textbf{A100}
& \textbf{H200}
& & \\
\hline
\textcolor{gray}{No Memory}
& \textemdash
& \textcolor{gray}{5.35}
& \textcolor{gray}{30.85}
& \textcolor{gray}{59.22}
& \textcolor{gray}{88.49}
& \textcolor{gray}{73.03} \\
\hline
Sliding Window
& \textemdash
& 3.22
& 28.21
& 56.08
& 87.09
& 69.70 \\
Flex~\cite{flex}
& 56M
& 4.66
& 26.99
& 54.01
& 87.14
& 70.11 \\
QT-Former~\cite{ORION}
& 77M
& \textbf{5.64}
& 28.01
& 54.02
& 87.37
& 71.06 \\
\hline
\rowcolor{blue!10}
\gls{RAM}
& \textbf{17M}
& 3.89
& \textbf{29.96}
& \textbf{56.71}
& \textbf{90.95}
& \textbf{77.27} \\
\bottomrule
\end{tabular}}
\end{minipage}
\vspace{-12pt}
\end{table*}

\textbf{Comparison with explicit action history.} To test whether past ego waypoints can substitute for latent action-token recurrence of \gls{RAM}, we augment the no-\gls{RAM} variant of FIVE-VLA with the previous 3 or 10 ego positions, encoded through an MLP (similar to target points) and supplied to the \gls{LLM}. \cref{tab:actionhistory} shows that adding waypoint history improves open-loop metrics (ADE/FDE) but significantly degrades closed-loop performances (DS/SR). However, the open-loop improvements do not establish improved driving as these errors are measured with recorded expert history. Specifically, this divergence between open- and closed-loop reflects the well-documented copycat shortcut and covariate shift problems~\cite{wen2020copycat,codevilla2019limitations,dehaan2019causal}: In open-loop, the model receives perfect expert history and minimises error by simply extrapolating past motion rather than understanding the visual scene. However, in closed-loop driving, the model must feed back its own imperfect predictions, causing small geometric errors to compound catastrophically. For example, Bench2Drive SR drops from 73.03\% to 64.39\% and 70.00\%, whereas \gls{RAM} reaches 77.27\%. \textit{In contrast, \gls{RAM} recurrently uses its own latent action tokens and improves closed-loop DS/SR over the no-memory baseline, without explicit past ego-waypoint inputs.}

\noindent \textbf{Comparison with vision-domain temporal baselines.} We replace \gls{RAM} with sliding window, Flex~\cite{flex} and QT-Former~\cite{ORION} in FIVE-VLA to investigate their potential as temporal memory mechanisms. As training with streams can degrade performance due to reduced intra-batch variability (see \cref{subsec:RAMfinetuning}), we evaluate these approaches under two regimes. First, incorporated from the second training stage with the same 4-epoch budget as \gls{RAM}, none of the vision-domain baselines produce gains over the no-memory baseline (Tab.~\ref{tab:othermemorytypes}). Unlike \gls{RAM}, these approaches depart from the pretrained \gls{VLM}'s design (sliding window exceeds pretraining vision-token counts, while Flex and QT-Former insert modules with randomly initialised queries between the vision encoder and \gls{LLM}), potentially hindering transfer learning. Second, incorporating them from the first training stage with gradient accumulation and up to 60 epochs of training yields marginal gains, yet still falls short of both the no-memory baseline and \gls{RAM} (results are included in App.~\ref{app:othermemory}). We argue that these departures from the pretrained \gls{VLM}'s design can disrupt its representations, making recovery difficult with the 140 hours of closed-loop training data available. In contrast, \textit{\gls{RAM} operates in the action domain, preserving the pretrained VLM while still capturing the temporal dependencies critical for driving, and achieves the best driving performance with only 17M parameter count (Tab.~\ref{tab:othermemorytypes})}.

\subsection{Ablation Analyses}
\label{subsec:ablation}

\begin{table*}[t]
\begin{minipage}{0.53\textwidth}
\centering
\setlength{\tabcolsep}{0.25em}
\caption{\textbf{Component ablations from SimLingo to FIVE-VLA.} After the backbone replacement, efficient driving and \gls{RAM} are tested separately and jointly. All modes omit \gls{CoT}.}
\vspace{-4pt}
\label{tab:incremental}
\scalebox{0.62}{
\begin{tabular}{l|ccc|ccc}
\toprule
\textbf{Model} & \textbf{Image Enc.} & \makecell{\textbf{Efficient}\\\textbf{Driving}} & \textbf{RAM} & \textbf{fps (T4)} $\uparrow$ & \textbf{DS} $\uparrow$ & \textbf{SR (\%)} $\uparrow$ \\
\midrule
SimLingo~\cite{SimLingo} & InternViT & \xmark & \xmark & 0.50 & 85.88 & 68.18 \\
\hline
\quad $\bullet$ FastVLM & FastViTHD & \xmark & \xmark & 2.20 & 88.49 & 73.03 \\
\quad $\bullet$ Efficient Driving & FastViTHD & \cmark & \xmark & \textbf{5.35} & 88.49 & 73.03 \\
\quad $\bullet$ RAM & FastViTHD & \xmark & \cmark & 1.46 & \textbf{90.95} & \textbf{77.27} \\
\hline
\rowcolor{blue!10}
FIVE-VLA & FastViTHD & \cmark & \cmark & 3.89 & \textbf{90.95} & \textbf{77.27} \\
\bottomrule
\end{tabular}}
\end{minipage}
\hfill
\begin{minipage}{0.46\textwidth}
\setlength{\tabcolsep}{1em}
\centering
\caption{\textbf{The effect of different \gls{RAM} inference modes.} \gls{RAM} accumulation outperforms bypassing \gls{RAM}, unboundedly using it or resetting.}
\label{tab:memoryaccumulation}
\scalebox{0.7}{
\begin{tabular}{cc|cc} 
\toprule
\textbf{\gls{RAM}}&\textbf{\gls{RAM} Inference Mode}&\textbf{DS} $\uparrow$&\textbf{SR(\%)} $\uparrow$ \\  \hline
\xmark&N/A&88.49&73.03\\  \hline 
\cmark&Bypass&88.23&72.42\\
\cmark&Unbounded&89.74&73.94\\ 
\cmark&Reset&90.21&75.91 \\ \hline
\rowcolor{blue!10} \cmark&Accumulation&\textbf{90.95}&\textbf{77.27}\\
\bottomrule
\end{tabular}}
\end{minipage}
\vspace{-12pt}
\end{table*}

\textbf{Contribution of Main Design Choices.} \cref{tab:incremental} evaluates the individual and combined contributions of FIVE-VLA's main components to efficiency and driving. As both FastVLM and SimLingo's backbone share the same Qwen2 \gls{LLM} architecture~\cite{qwen2}, one key difference lies in the image encoder. Using FastViTHD instead of InternViT~\cite{miniinternvl} yields a substantial gain of 4.85 SR and a $\sim4\times$ speed-up on T4, confirming the importance of FastViTHD, which processes high-resolution images without tile-based partitioning. Adding efficient driving mode brings a further $\sim 2\times$ speed-up at no cost to driving performance, as it eliminates \gls{AR} text generation. \gls{RAM} also provides substantial gains (4.24 SR, 2.46 DS) with only a modest efficiency cost. Combining all three, FIVE-VLA reaches 77.27\% SR and 3.89 fps on T4, a 9.09 points SR gain and $\sim$$8\times$ speed-up over SimLingo without \gls{CoT}. \textit{Notably, efficient driving mode and \gls{RAM} address orthogonal bottlenecks as the former reduces the latency while the latter captures temporal dependencies, making them complementary.}

\textbf{\gls{RAM} inference modes.} To isolate the benefit of accumulation, we evaluate the same trained FIVE-VLA model under four distinct \gls{RAM} inference modes in \cref{tab:memoryaccumulation}. First, we use a \textit{bypass} strategy where action tokens are passed directly to the \gls{LLM} at each time step, effectively disabling \gls{RAM}. As expected, this yields 88.23 \gls{DS}, similar to the model trained without \gls{RAM} (88.49 \gls{DS}), confirming again that the gains are attributable to \gls{RAM}. Second, we test using \gls{RAM} in an \textit{unbounded} manner that continuously aggregates memory without considering the training stream length. While this (89.74 \gls{DS}) outperforms bypassing \gls{RAM}, it remains suboptimal, likely due to the inference streaming length exceeding the training one. In contrast, \textit{reset} and \textit{accumulation}, the strategies aligned with training, yield gains of around 2 points in \gls{DS} and 3-4 points in \gls{SR} over the baseline without \gls{RAM}, where accumulation performs the best. 

\section{Conclusions}
\label{sec:conclusion}
We presented FIVE-VLA, an efficient and effective \gls{VLA} for \gls{AD} that establishes a new \gls{SOTA} on Bench2Drive and Fail2Drive. FIVE-VLA addresses three critical limitations of existing \glspl{VLA}: it processes $448\times896$ images with only 98 vision tokens via FastVLM, bypasses text generation entirely, and introduces \gls{RAM}, a lightweight memory module conditioning action prediction on previous action tokens without modifying the backbone \gls{VLM}. With only 641M parameters, FIVE-VLA outperforms the previous \gls{SOTA} \gls{VLA} by an absolute margin of 10\% in success rate and runs $8\times$ faster on edge-device GPUs. Experiments on the NVIDIA Physical AI AV dataset further demonstrate that FIVE-VLA's benefits extend beyond simulation to real-world and multi-view settings.

}

\begin{ack}
We gratefully acknowledge Hamza Adnan for his contributions to adapting NAVSIM-style non-reactive simulation, Alexey Zakharov for his thoughtful feedback and constructive suggestions on the manuscript, Anthony Knittel for stimulating technical discussions and valuable exchanges of ideas throughout the project, and Alexander Bartler for data preparation and support. We also thank our other colleagues at Bosch (including FIVE AI) who had direct or indirect support in helping us make faster progress.
\end{ack}

\blockcomment{
\section*{References}

References follow the acknowledgments in the camera-ready paper. Use unnumbered first-level heading for
the references. Any choice of citation style is acceptable as long as you are
consistent. It is permissible to reduce the font size to \verb+small+ (9 point)
when listing the references.
Note that the Reference section does not count towards the page limit.
\medskip
}

\bibliographystyle{unsrtnat}
\bibliography{main}

\clearpage

\section*{APPENDICES}
\begingroup
\small 
\tableofcontents
\endgroup
\clearpage

\renewcommand{\thefigure}{A.\arabic{figure}}
\renewcommand{\thetable}{A.\arabic{table}}
\renewcommand{\theequation}{A.\arabic{equation}}
\renewcommand{\thealgorithm}{A.\arabic{algorithm}}

\renewcommand{\thesection}{A}
\section{Further Details and Experiments}
\label{app:experiments}

In this section, we present further details on the NVIDIA real-world and multi-view evaluation, followed by additional experiments probing the robustness of \gls{RAM} to corrupted memory and different inference camera frequencies. We then provide extended discussions and results on alternative vision-based memory mechanisms, supplementary closed-loop evaluations on Bench2Drive and Fail2Drive, and an analysis of the impact of \gls{AR} text generation on throughput. Finally, we showcase the language capabilities and \gls{VQA} performance of FIVE-VLA and discuss the main limitations of our work.

\subsection{Training Protocol}
\label{app:trainingprotocol}
Unless otherwise specified, closed-loop FIVE-VLA is trained for six epochs without \gls{RAM}, followed by four epochs of \gls{RAM} fine-tuning. On NVIDIA Physical AI AV, SimLingo and FIVE-VLA without \gls{RAM} are trained for one epoch; single- and four-view \gls{RAM} models initialise from their respective no-memory checkpoints at epoch 0.5 and fine-tune for two epochs. Extending training without \gls{RAM} does not recover the gains from \gls{RAM} fine-tuning in either setting (\cref{tab:nvidia_training_duration,tab:longertraining}). For both open- and closed-loop experiments, \gls{RAM} fine-tuning uses streams of $T=4$ observations.

\subsection{Open-Loop Evaluation Details and Further Experimental Results on NVIDIA Dataset}
\label{app:nvidiaevaluation}

We evaluate the open-loop models on held-out NVIDIA Physical AI AV clips using two complementary metric families: displacement errors on the directly predicted waypoints, and NAVSIM-style~\cite{NAVSIM} scores on controller-executed trajectories. The following details apply to both single-view and four-view models. 

\subsubsection{Displacement Metrics}
Let $\mathbf{O}_q$ denote the predicted speed or path sequence ($\SpeedWaypoints$ or $\PathWaypoints$; \cref{subsec:overview}), with $q\in\{\mathrm{speed},\mathrm{path}\}$, and $\mathbf{O}_q^*$ its corresponding expert targets. Writing $[\cdot]_i$ for the $i$-th waypoint, average displacement error (ADE) and final displacement error (FDE) are
\begin{align}
    \mathrm{ADE}_{q} &= \frac{1}{H_q}\sum_{i=1}^{H_q}\bigl\lVert[\mathbf{O}_{q}]_i-[\mathbf{O}_{q}^{*}]_i\bigr\rVert_2, \nonumber\\
    \mathrm{FDE}_{q} &= \bigl\lVert[\mathbf{O}_{q}]_{H_q}-[\mathbf{O}_{q}^{*}]_{H_q}\bigr\rVert_2.
    \label{eq:nvidia_displacement}
\end{align}
For speed waypoints, $H_{\mathrm{speed}}=30$: waypoint $i$ is compared with the expert position at future time $0.1i$ seconds. These errors measure time-aligned position accuracy, not scalar speed error.

For path waypoints, we construct the ground truth from the expert driving by using entire future. We connect consecutive positions to form a polyline, whose arc length is the cumulative horizontal distance between them. We estimate the current ego's arc-length coordinate $s_0$ by local projection onto this polyline, then linearly interpolate the recorded positions at $s_0+i$ metres for $i=1,\ldots,30$. Transforming these positions into the current ego frame gives the ground-truth path waypoints $[\PathWaypoints^{*}]_i$. Each predicted waypoint $[\PathWaypoints]_i$ is compared directly with its corresponding interpolated target ($H_{\mathrm{path}}=30$). Path FDE therefore uses the target 30 metres ahead along the recorded path.

All errors are in metres and compare direct model predictions with expert targets, not the simulated ego rollout. A frame contributes only when its recorded ground truth covers the full horizon of the corresponding metric. In particular, frames with less than 30 metres of remaining recorded path are excluded from path ADE/FDE rather than scored against padded endpoints. We average each metric over its valid frames, equivalently weighting clip-level means by that metric's valid-frame count.

\subsubsection{NAVSIM-Style Non-Reactive Simulation}
\label{app:nvidianavsim}
We adapt the NAVSIM~\cite{NAVSIM} scoring approach to native NVIDIA clips. At each scored observation, the open-loop evaluator reuses the predicted path and speed waypoints to simulate a 3-second ego rollout. Surrounding traffic follows recorded trajectories and does not react to the ego. The next model input remains the next recorded observation, not a simulated observation; the simulated ego and controller state are reinitialised for each scored observation. 

\paragraph{Ego rollout and controller.}
The ego starts at the origin of the current ego frame, with zero relative heading and its recorded speed. A kinematic bicycle model follows the predicted path using pure-pursuit steering referenced to the rear axle, with adaptive lookahead $\max(2\,\mathrm{m},(0.2\,\mathrm{s})v)$ for current speed $v$. The controller selects the first usable waypoint meeting the lookahead distance, or the farthest usable waypoint when none reaches it, and floors the steering denominator at the minimum lookahead distance. If the path cannot provide a valid lookahead point, steering falls back to the speed-waypoint geometry. Longitudinal target speeds are obtained from consecutive speed-waypoint displacements divided by 0.1 seconds, including the displacement from the origin to the first waypoint. A PID controller uses gains $(K_P,K_I,K_D)=(5.0,0.01,0.03)$ and a five-step integral window. Each 0.1-second waypoint interval is integrated using five 0.02-second substeps: longitudinal control updates at each substep, while steering is held fixed within the interval. The resulting 30 simulated poses are scored at offsets $0.1,0.2,\ldots,3.0$ seconds.

\paragraph{Vehicle geometry and recorded traffic.}
Ego length, width, and wheelbase are read from the clip's vehicle metadata; the bounding-box centre offset from the rear axle approximates the centre-of-gravity location for the bicycle model. Clips lacking the required vehicle metadata are skipped. All available obstacle classes are retained. Recorded obstacle boxes are transformed from their capture-time rig frames into the fixed ego frame of the scored observation using recorded egomotion, then interpolated to the simulation timestamps. Future boxes are aligned with the simulated future poses, while boxes at the initial observation are used to exclude tracks already overlapping the ego before the rollout. Missing obstacle labels, missing egomotion, or a scored timestamp outside the obstacle-label coverage exclude NC and TTC for that observation.

\paragraph{No at-fault collisions (NC).}
Our adapted NC is binary: $\mathrm{NC}_i\in\{0,1\}$ for each scored observation $i$; there is no $0.5$ penalty. At each rollout step, the planar oriented ego box is tested for intersection with the recorded obstacle boxes. A contact with an obstacle ahead of the ego (its centre lies in the ego's forward half-plane) or stopped (speed below $0.005\,\mathrm{m/s}$) sets NC to zero, irrespective of obstacle class; otherwise NC is one. Moving lateral/rear contacts are not classified as at-fault, and subsequent contacts with those tracks are ignored within the rollout. Tracks already overlapping the ego at the initial observation are excluded. Thus, NC$=1$ means no qualifying at-fault contact, not necessarily no contact; this rule approximates rather than fully determines collision responsibility.

\paragraph{Time-to-collision compliance (TTC).}
At each eligible rollout step, the ego box is projected at its current speed and heading, and tested against the recorded future obstacle boxes. The nominal one-second lookahead is sampled at offsets $0,0.3,0.6,0.9$ seconds. A projected intersection with an obstacle ahead of the current ego pose gives a score of zero; otherwise the score is one. Steps with ego speed below $0.005\,\mathrm{m/s}$ are skipped. 

\paragraph{NC and TTC violation percentages.}
For each scored observation $i$, $\mathrm{NC}_i,\mathrm{TTC}_i\in\{0,1\}$ indicate whether its 3-second rollout is free of the respective violations. Let $\mathcal{V}_{\mathrm{NC}}$ and $\mathcal{V}_{\mathrm{TTC}}$ denote the sets of observations with valid annotations for the respective metrics. We report their complementary violation percentages as
\begin{align}
    \mathrm{NC\text{-}V} &= \frac{100}{|\mathcal{V}_{\mathrm{NC}}|}\sum_{i\in\mathcal{V}_{\mathrm{NC}}}(1-\mathrm{NC}_i)=100(1-\mathrm{NC}), \nonumber\\
    \mathrm{TTC\text{-}V} &= \frac{100}{|\mathcal{V}_{\mathrm{TTC}}|}\sum_{i\in\mathcal{V}_{\mathrm{TTC}}}(1-\mathrm{TTC}_i)=100(1-\mathrm{TTC}).
    \label{eq:nvidia_violation_percentages}
\end{align}
Here, $\mathrm{NC}$ and $\mathrm{TTC}$ denote the unscaled dataset-level means in $[0,1]$. As in the main paper, \emph{NC-V (\%)} is the at-fault collision percentage under the attribution rule above, and \emph{TTC-V (\%)} is the TTC-violation percentage; both are lower-is-better. Each denominator counts valid evaluated observations, with unavailable observations excluded rather than treated as violation-free; an empty valid set yields an unavailable metric, not zero. These percentages count observations whose rollouts contain at least one qualifying violation, not unique collision events, and overlapping rollout horizons can refer to the same recorded traffic interaction.

\paragraph{Ego progress (EP).}
The recorded future ego trajectory over the 3-second horizon defines the reference polyline. We project the first and last simulated positions onto this polyline, divide their non-negative arc-length difference by the reference polyline length, and clip the ratio to $[0,1]$. 

\paragraph{Comfort (C).}
Comfort is a binary check on longitudinal acceleration, lateral acceleration, and yaw rate. For the 30-sample rollout, speed and unwrapped heading are smoothed using a nine-sample, second-order Savitzky--Golay filter; kinematic derivatives are estimated with the same filter, excluding four samples at each boundary. The thresholds are $a_{\mathrm{long}}\in[-4.05,2.40]\,\mathrm{m/s^2}$, $|a_{\mathrm{lat}}|\leq4.89\,\mathrm{m/s^2}$, and $|\dot{\psi}|\leq0.95\,\mathrm{rad/s}$. The upper and lower longitudinal limits, lateral acceleration, and yaw rate are checked separately, each permitting at most one violating sample; yaw rate is checked only above $0.5\,\mathrm{m/s}$. Jerk and yaw acceleration are not evaluated because their numerical estimates are unreliable over these short trajectories. 

\paragraph{Adapted PDM score (PDMS) and aggregation.}
Drivable-area compliance (DAC) is omitted because the required map annotations are unavailable in this evaluation. When all remaining components are available, the per-observation score is
\begin{align}
    \mathrm{PDMS} = \mathrm{NC}\,\frac{5\,\mathrm{EP}+5\,\mathrm{TTC}+2\,\mathrm{C}}{12}.
    \label{eq:nvidia_pdms}
\end{align}
All components in \cref{eq:nvidia_pdms} are unscaled per-observation scores in $[0,1]$; a qualifying at-fault collision therefore sets PDMS to zero. Unavailable components are omitted: additive weights are renormalised over the available components, and the NC multiplicative gate is omitted when NC cannot be computed. Without NC and TTC, the aggregate reflects only progress and comfort, not safety compliance. Each submetric is averaged over its valid observations; PDMS is computed per observation before averaging, rather than reconstructed from the reported mean submetrics. The supplementary five-score suite reports NC, EP, TTC, comfort, and PDMS as $100$ times their respective means, all higher-is-better. Its NC and TTC columns are compliance percentages, equal to $100-\mathrm{NC\text{-}V}$ and $100-\mathrm{TTC\text{-}V}$ before rounding. This adapted suite omits map-based compliance and modifies collision attribution and comfort checks, so it is not the official NAVSIM benchmark score.

\subsubsection{Further Results on NVIDIA Physical AI AV}

\paragraph{NAVSIM-style evaluation results.}
\label{app:nvidiaresults}
\cref{tab:nvidiaav_app} reports the main-paper safety measures alongside the five-score NAVSIM-style suite used in our adaptation. NC/TTC and NC-V/TTC-V describe the same outcomes, not independent evidence; violation percentages make differences between near-ceiling compliance scores easier to read. FIVE-VLA with \gls{RAM} achieves the highest NC, EP, TTC, and adapted PDMS in both view settings. For single-view FIVE-VLA, adding \gls{RAM} increases EP from 95.48 to 96.02 and PDMS from 97.03 to 97.32: improved collision and TTC compliance accompanies greater progress along the recorded expert trajectory. Four-view FIVE-VLA attains the highest EP (96.06) and PDMS (97.35), although the gains over its single-view counterpart are small. Comfort is not uniformly improved: it decreases from 99.45 to 99.39 with single-view \gls{RAM}, and four-view FIVE-VLA scores 99.32 versus SimLingo's 99.33. Thus, the suite indicates a better progress--safety balance with small comfort trade-offs under this non-reactive protocol, not a guarantee of reactive closed-loop safety.

\begin{table}[t]
\centering
\setlength{\tabcolsep}{0.4em}
\caption{\textbf{NAVSIM-style evaluation on NVIDIA Physical AI AV.} Left: NC-V/TTC-V violation percentages, as in \cref{tab:nvidiaav}. Right: the five-score suite, with mean scores scaled to 0--100; NC/TTC report compliance, and PDMS uses the adapted formula in \cref{eq:nvidia_pdms}. FIVE-VLA includes \gls{RAM} unless noted otherwise. \textbf{Bold}: best within each camera setting.}
\label{tab:nvidiaav_app}
\resizebox{\linewidth}{!}{%
\begin{tabular}{@{}lc|cc|ccccc@{}}
\toprule
& & \multicolumn{2}{c|}{\textbf{Safety violations (\%)}} & \multicolumn{5}{c}{\textbf{NAVSIM-style scores (0--100)}} \\
\cmidrule(lr){3-4}\cmidrule(lr){5-9}
\textbf{Method} & \textbf{Cameras} & \textbf{NC-V} $\downarrow$ & \textbf{TTC-V} $\downarrow$ & \textbf{NC} $\uparrow$ & \textbf{EP} $\uparrow$ & \textbf{TTC} $\uparrow$ & \textbf{Comfort} $\uparrow$ & \textbf{PDMS} $\uparrow$ \\
\midrule
SimLingo~\cite{SimLingo} & 1 & 0.87 & 1.26 & 99.13 & 95.55 & 98.74 & 99.37 & 97.02 \\
FIVE-VLA (w/o \gls{RAM}) & 1 & 0.84 & 1.25 & 99.16 & 95.48 & 98.75 & \textbf{99.45} & 97.03 \\
\rowcolor{blue!10}
FIVE-VLA & 1 & \textbf{0.78} & \textbf{1.16} & \textbf{99.22} & \textbf{96.02} & \textbf{98.84} & 99.39 & \textbf{97.32} \\
\midrule
SimLingo~\cite{SimLingo} & 4 & 0.82 & 1.21 & 99.18 & 95.79 & 98.79 & \textbf{99.33} & 97.17 \\
\rowcolor{blue!10}
FIVE-VLA & 4 & \textbf{0.75} & \textbf{1.15} & \textbf{99.25} & \textbf{96.06} & \textbf{98.85} & 99.32 & \textbf{97.35} \\
\bottomrule
\end{tabular}}
\end{table}

\paragraph{Effect of Training Duration.}
\label{app:nvidiatrainingduration}
We train models without \gls{RAM} on NVIDIA dataset for 1 epoch, as further training results in a worse performance in our experiments. We compare single-view FIVE-VLA without \gls{RAM} trained for one and two epochs to test whether extending no-memory training recovers the gains from recurrent memory. Separately, we compare one and two epochs of \gls{RAM} fine-tuning, both initialised from the same no-memory checkpoint at epoch 0.5. \Cref{tab:nvidia_training_duration} shows that extending no-memory training from one to two epochs increases speed ADE/FDE from 0.549/1.464 to 0.570/1.526\,m (3.9\%/4.2\%), rather than recovering the memory models' accuracy. Extending \gls{RAM} fine-tuning from one to two epochs reduces ADE/FDE from 0.412/1.150 to 0.402/1.128\,m (2.3\%/1.9\%) and slightly improves path errors and violation percentages.

\begin{table}[t]
\centering
\setlength{\tabcolsep}{0.5em}
\caption{\textbf{Single-view training-duration ablation on NVIDIA Physical AI AV.} Epochs count base training without \gls{RAM}, or fresh \gls{RAM} fine-tuning from the no-memory epoch-0.5 checkpoint. Errors are in metres; NC-V/TTC-V are percentages; EP/C/PDMS are on a 0--100 scale. \textbf{Bold}: best at displayed precision.}
\label{tab:nvidia_training_duration}
\resizebox{\linewidth}{!}{%
\begin{tabular}{cc|cc|cc|ccccc}
\toprule
& & \multicolumn{2}{c|}{\textbf{Speed (3\,s)}} & \multicolumn{2}{c|}{\textbf{Path (30\,m)}} & \multicolumn{5}{c}{\textbf{Non-reactive evaluation}} \\
\cmidrule(lr){3-4}\cmidrule(lr){5-6}\cmidrule(lr){7-11}
\textbf{\gls{RAM}} & \textbf{Epochs} & \textbf{ADE} $\downarrow$ & \textbf{FDE} $\downarrow$ & \textbf{ADE} $\downarrow$ & \textbf{FDE} $\downarrow$ & \textbf{NC-V} $\downarrow$ & \textbf{TTC-V} $\downarrow$ & \textbf{EP} $\uparrow$ & \textbf{C} $\uparrow$ & \textbf{PDMS} $\uparrow$ \\
\midrule
\xmark & 1 & 0.549 & 1.464 & 0.213 & 0.405 & 0.84 & 1.25 & 95.48 & \textbf{99.45} & 97.03 \\
\xmark & 2 & 0.570 & 1.526 & 0.214 & 0.406 & 0.86 & 1.26 & 95.55 & \textbf{99.45} & 97.05 \\
\midrule
\cmark & 1 & 0.412 & 1.150 & 0.211 & 0.400 & 0.79 & 1.17 & \textbf{96.03} & 99.35 & 97.31 \\
\rowcolor{blue!10}
\cmark & 2 & \textbf{0.402} & \textbf{1.128} & \textbf{0.207} & \textbf{0.392} & \textbf{0.78} & \textbf{1.16} & 96.02 & 99.39 & \textbf{97.32} \\
\bottomrule
\end{tabular}}
\end{table}

\subsection{Robustness of \gls{RAM}}
\label{app:ramrobustness}
Here, we discuss the robustness of \gls{RAM} in open-loop and closed-loop experiments.
\subsubsection{Effect of Perturbations on RAM}

\textbf{Protocol and reference conditions.} \gls{RAM} is a streaming memory: at observation $t$, the previous action-token array $\RAMInput^{t-1}$ conditions the current trajectory prediction. We probe its sensitivity by corrupting the memory read before the memory encoder, at inference only, on a fixed 10\% subset of NVIDIA test clips (approximately 35 hours, open-loop). All conditions use the same checkpoint and clip subset, without retraining. \emph{Clean} uses unperturbed \gls{RAM} accumulation; \emph{Ignore-RAM} bypasses memory conditioning in the same checkpoint, rather than substituting a separately trained no-memory model. Ignore-RAM is a reference for the benefit of using memory, not an upper bound on error under arbitrary corruption. We report the displacement metrics defined in \cref{app:nvidiaevaluation}.

\textbf{Perturbations.} We apply each perturbation separately to the retrieved array before it enters the memory encoder.
\begin{itemize}
    \item \textbf{Noise:} $\RAMInput^{t-1}\mapsto\RAMInput^{t-1}+\sigma r_{t-1}\boldsymbol{\epsilon}$, where the entries of $\boldsymbol{\epsilon}$ are independent standard Gaussian variables and $r_{t-1}=\max(\operatorname{RMS}(\RAMInput^{t-1}),10^{-6})$. RMS is computed over the retrieved tensor's elements, so $\sigma\in\{0.5,1.0,2.0\}$ scales noise relative to feature magnitude rather than specifying an absolute embedding-space deviation.
    \item \textbf{Wrong action:} With probability $p\in\{0.25,0.5\}$, replace the retrieved array with a model-generated memory from a fixed bank built by farthest-point sampling in standardised action-descriptor space. Each candidate is described by the two-dimensional endpoint $(x_{3\mathrm{s}},y_{3\mathrm{s}})$ of its predicted speed-waypoint trajectory, encoding forward progress and lateral displacement, rather than a velocity vector. Candidates are ranked by Euclidean distance from the previous prediction's descriptor after per-coordinate standardisation; a replacement is sampled uniformly from the eight most distant candidates. The bank remains fixed during evaluation, and replacement is applied only when a compatible state and previous descriptor are available.
    \item \textbf{Dropped frame:} With probability $p\in\{0.25,0.5\}$, retrieve the preceding cached memory, nominally $\RAMInput^{t-2}$ instead of $\RAMInput^{t-1}$, when a shape-compatible cache is available. This models a stale memory read: it does not discard the current image or skip the memory write, and it does not accumulate an arbitrarily long delay across consecutive perturbations.
\end{itemize}

\begin{table}[t]
\centering
\small
\setlength{\tabcolsep}{0.6em}
\caption{\textbf{Inference-time robustness of \gls{RAM} on a fixed 10\% NVIDIA test subset.} All errors are in metres; lower is better. Ignore-RAM disables memory conditioning in the same checkpoint. The clean subset result is distinct from the full-test result in \cref{tab:nvidiaav}.}
\label{tab:ram_robustness}
\begin{tabular}{lcc|cc}
\toprule
& \multicolumn{2}{c|}{\textbf{Trajectory (3\,s)}} & \multicolumn{2}{c}{\textbf{Path (30\,m)}} \\
\textbf{Condition} & \textbf{ADE} $\downarrow$ & \textbf{FDE} $\downarrow$ & \textbf{ADE} $\downarrow$ & \textbf{FDE} $\downarrow$ \\
\midrule
\rowcolor{blue!10}
Clean & 0.415 & 1.153 & 0.220 & 0.416 \\
Ignore-RAM & 0.548 & 1.454 & 0.219 & 0.410 \\
\midrule
Noise, $\sigma=0.5$ & 0.432 & 1.188 & 0.216 & 0.410 \\
Noise, $\sigma=1.0$ & 0.497 & 1.324 & 0.213 & 0.401 \\
Noise, $\sigma=2.0$ & 0.612 & 1.565 & 0.219 & 0.414 \\
Wrong action, $p=0.25$ & 0.741 & 1.778 & 0.238 & 0.465 \\
Wrong action, $p=0.5$ & 1.015 & 2.293 & 0.254 & 0.500 \\
Dropped frame, $p=0.25$ & 0.430 & 1.181 & 0.216 & 0.412 \\
Dropped frame, $p=0.5$ & 0.448 & 1.215 & 0.214 & 0.400 \\
\bottomrule
\end{tabular}
\end{table}

\textbf{Results and limitations.} In \cref{tab:ram_robustness}, ignoring \gls{RAM} raises trajectory ADE by approximately 32\% and FDE by 26\% relative to Clean. Moderate noise ($\sigma\leq1$) and stale-memory reads retain lower trajectory errors than Ignore-RAM; at $p=0.5$, stale reads increase ADE by approximately 8\%. This tolerance is not universal: stronger noise ($\sigma=2$) exceeds the Ignore-RAM reference, and wrong-action injection is the most damaging stressor, raising ADE by approximately 145\% at $p=0.5$. Path errors remain close to Clean under noise and stale reads, but increase from 0.220/0.416 to 0.254/0.500 under the strongest wrong-action condition. The larger changes in time-parameterised trajectory errors are consistent with sensitivity in motion timing and longitudinal dynamics.

\begin{table}[t]
\centering
\caption{\textbf{The effect of the difference in training and evaluation camera frequency.} Fps-aligned Reset ensures that during inference, \gls{RAM} uses historical action tokens with the same frequency of the training. Fps-aligned Reset performs similar to Reset.}
\setlength{\tabcolsep}{2.5em}
\label{tab:fpsapp}
\scalebox{0.90}{
\begin{tabular}{cc|cc} 
\toprule
\textbf{\gls{RAM}}&\textbf{\gls{RAM} Inference Mode}&\textbf{DS} $\uparrow$&\textbf{SR(\%)} $\uparrow$ \\ \midrule
\xmark&N/A&88.49&73.03\\ \hline 
\cmark&fps-aligned Reset&90.02&\textbf{76.21}\\
\cmark&Reset&\textbf{90.21}&75.91 \\ 
\bottomrule
\end{tabular}}
\end{table}

\subsubsection{The Effect of Different Camera Frequency on RAM}
Bench2Drive closed-loop evaluation provides images to the models at 20 fps by default, while SimLingo driving dataset we use for training is collected at 4fps. Considering that \gls{RAM} is a memory module, which should provide useful information about the temporal dynamics, here we investigate whether this difference in the sensor frequency between training and evaluation affects the driving capabilities of FIVE-VLA.

To investigate this, we design a \gls{RAM} inference mode, called \textit{fps-aligned reset}, as a \gls{RAM} reset variant that enables \gls{RAM} to process information at 4 fps, matching the training camera frequency. Specifically, we maintain a queue of previous action tokens $[\RAMInput^{t-5}, \RAMInput^{t-4}$, $\RAMInput^{t-3}$, $\RAMInput^{t-2}$, $\RAMInput^{t-1}]$, and provide $\RAMInput^{t-5}$ to \gls{RAM} at the current time step $t$, accounting for the 5$\times$ difference in camera frequency between training (4 fps) and inference (20 fps).

We evaluate FIVE-VLA with fps-aligned reset and compare it with standard reset in \cref{tab:fpsapp}. The results show that fps-aligned reset performs very similarly to \gls{RAM} reset, demonstrating the model's robustness to this frequency difference. As motivated in \cref{sec:introduction}, this observation is aligned with its similarity to TRM. As the scenes evolve slowly, similar to TRM, \gls{RAM} decreases the reliance on inferring actions from scratch at each time step, resulting in more effective utilisation of the \gls{LLM}. Having learned to benefit from the previous frames,  FIVE-VLA becomes robust to this frequency difference.

\subsection{Further Details and Discussions on Other Memory Mechanisms}
\label{app:othermemory}

In this section, we elaborate on our implementation details on other memory mechanisms we compared with. We also present additional experiments for sliding window, Flex, and QT-Former, where we trained these models for an extended duration from the first stage of training.

\subsubsection{Implementation Details on Action-History Baselines}
\label{app:actionhistory}
\textbf{Input construction.} The explicit action-history baselines in \cref{tab:actionhistory} use FIVE-VLA without \gls{RAM}, retaining a single current image, current ego speed, and navigational inputs. We add $L\in\{3,10\}$ past two-dimensional ego positions in metres, transformed into the current ego frame and ordered from oldest to newest, excluding the current origin. Positions are sampled at 4\,Hz for the closed-loop setting and 10\,Hz for the open-loop setting. A dedicated MLP, with dimensions $2\rightarrow256\rightarrow512\rightarrow896$ and ReLU activations between linear layers, maps each position to one \gls{LLM} input embedding. The resulting $L$ history tokens are inserted after the speed input and before navigation, using the same embedding-insertion mechanism as target points but separate MLP weights.

\textbf{Training and evaluation histories.} These baselines are trained without a \gls{RAM} fine-tuning stage. Training and open-loop evaluation use recorded ego-pose histories, excluding samples without a complete history. Closed-loop evaluation instead buffers online ego-pose estimates from the model-controlled vehicle within each route. Thus, these inputs describe past realised ego motion, not stored arrays of previously predicted future waypoints or steering/throttle/brake commands. In contrast, \gls{RAM} recurrently conditions on its own previous latent action tokens in both evaluation settings.

\subsubsection{Implementation details on Vision-based Temporal Baselines}
We train all approaches with a stream length of $T=4$. The \textit{Sliding window} approach tokenises each image separately using the vision encoder before feeding all tokens to \gls{LLM}. As our image encoder outputs 98 tokens per image, this results in 392 tokens being propagated to the \gls{LLM}. For sliding window and Flex, we include two implementations: (i) the standard one, where all images are fed through the entire model, and (ii) an efficient one, where we cache previous vision representations and propagate only the image at the last time step to the vision encoder. These implementations produce identical performance but differ in efficiency only. We used the efficient one while reporting the results in the main paper.

For \textit{Flex}~\cite{flex}, we implement both vanilla Flex and the variant of Flex with interleaving, where the model receives supervision at intermediate time steps. We use a compression rate of 5, as it is one of the tested settings in \cite{alpamayor1,flex} and we do not require very large compression given that our vision encoder already generates relatively few tokens. Specifically, this results in 20 tokens per image (down from 98 tokens per image). With 4 images, the total number of learnable scene queries is 80. We add time step encodings to the images. The Flex module, consisting of 8 self-attention-based transformer blocks, receives 472 tokens (80 learnable scene tokens and 392 image tokens) as input, drops the image tokens at its output, and propagates only the scene tokens to the \gls{LLM}. When training Flex with interleaving, we follow the description in the original paper, where we set the attention mask such that the $i$-th time step only uses the first $(i + 1) \times 20$ scene tokens where $i \in \{0,1,2,3\}$.

For \textit{QT-Former}, we similarly use a compression rate of 5 and introduce 16 history queries following~\cite{ORION}. We add time step encodings to the vision representations in the memory bank. To match the stream length of 4, the memory bank stores representations from the previous 3 time steps. To analyse all approaches under similar supervision settings, we do not use perception supervision for QT-Former. Accordingly, only 36 tokens (20 scene and 16 history tokens) are fed to the \gls{LLM} from the QT-Former.

For all vision-based memory approaches, we account for the difference between training camera frequency (4 fps) and evaluation camera frequency (20 fps) by employing a queue, through which we propagate historical images during inference following the training camera frequency (4 fps) to avoid potential domain shift.

\begin{table}[t]
\centering
\caption{\textbf{Comparison with other memory mechanisms trained from initialisation.} We train FIVE-VLA by replacing \gls{RAM} with alternative memory mechanisms, incorporating them from the beginning of training rather than in the second stage. The best configuration for each memory mechanism is \underline{underlined}. For two-stage training results, see \cref{tab:othermemorytypes}. Grad. Acc.: Gradient accumulation.}
\label{tab:memoryaccumulationapp}
\scalebox{0.88}{
\begin{tabular}{ccccc|cc} 
\toprule
\#&\textbf{Method}&\textbf{Training Strategy}&\textbf{\# Epochs}&\textbf{Grad. Acc. Steps}&\textbf{DS} $\uparrow$ & \textbf{SR(\%)} $\uparrow$\\ \hline
1&No Memory&Finetune FastVLM&6&-&88.49&73.03\\ \hline
2&Sliding Window&Finetune FastVLM&6&1&74.21&46.21 \\
3&Sliding Window&Finetune FastVLM&12&2&84.01&60.91 \\
4&Sliding Window&Finetune FastVLM&20&2&83.73&64.24\\
5&Sliding Window&Finetune FastVLM&40&4&84.37&66.52 \\
6&Sliding Window&Finetune FastVLM&30&2&86.75&70.91 \\
7&Sliding Window&Finetune FastVLM&60&4&\underline{86.76}&\underline{71.06}\\ 
8&Sliding Window&Finetune FastVLM&40&2&85.12&67.73\\ 
\hline 
9&QT-Former&Finetune FastVLM&12&2&82.47&63.94\\ 
10&QT-Former&Finetune FastVLM&20&2&86.20&68.64\\
11&QT-Former&Finetune FastVLM&30&2&\underline{87.21}&\underline{69.70} \\
12&QT-Former&Finetune FastVLM&40&2&85.90&69.55 \\
\hline
13&Flex&Finetune FastVLM&12&2&75.62&51.52 \\
14&Flex&Finetune FastVLM&30&2&\underline{85.30}&\underline{68.74} \\
15&Flex with interleaving&Finetune FastVLM&30&2&81.33&58.03 \\
16&Flex&Finetune FastVLM&40&2&84.30&65.80  \\
17&Flex with interleaving&Finetune FastVLM&40&2&81.85&61.52\\
\hline 
\rowcolor{blue!10} 18&\gls{RAM} Reset&Finetune \#1&4&4&90.21&75.91 \\
\rowcolor{blue!10} 19&\gls{RAM} Accumulation&Finetune \#1&4&4&\textbf{90.95}&\textbf{77.27}\\
\bottomrule
\end{tabular}}
\end{table}

\subsubsection{Further Experiments and Discussions}
Unlike the experiments in \cref{sec:experiments}, where memory mechanisms are incorporated in the second training stage (similar to \gls{RAM}), here we integrate them from the beginning to assess their performance in this configuration. We replace \gls{RAM} with alternative memory mechanisms in FIVE-VLA for these experiments. Given that model performance can degrade due to temporally-correlated data (as discussed in \cref{subsec:RAMfinetuning}), we employ gradient accumulation during training. Lines 2--17 in \cref{tab:memoryaccumulationapp} present results for different memory mechanisms.

\paragraph{The necessity for longer training.} As the first set of experiments, we train models for 12 epochs with gradient accumulation of 2, matching the number of updates in the baseline model (trained for 6 epochs without gradient accumulation). However, all alternative memory mechanisms perform poorly under this regime. Sliding window, QT-Former, and Flex achieve only 61\%, 64\%, and 52\% success rates, respectively (L3, L9, L13 in \cref{tab:memoryaccumulationapp}). Gradually increasing training epochs improves the success rate nearly to 71\%, 70\% and 69\%  for sliding window,  QT-Former and Flex respectively, approaching the baseline performance (around 73\% in L1). This shows that incorporating these memory mechanisms from initialisation requires substantially more training updates than the baseline (L1) as well as fine-tuning \gls{RAM} (L18, L19).

\paragraph{Impact of gradient accumulation.}
We investigate the effect of increasing gradient accumulation from 2 to 4, focusing on the sliding window approach due to its simplicity. To maintain comparable training updates, we proportionally increase the number of epochs. We test two regimes: (i) a shorter regime comparing 20 epochs with gradient accumulation of 2 versus 40 epochs with gradient accumulation of 4 (L4 vs. L5); and (ii) a longer regime comparing 30 versus 60 epochs (L6 vs. L7). While higher gradient accumulation benefits the shorter training regime, the improvement is negligible for the longer regime. For a more substantial effect on gradient accumulation, please compare L2 vs L3, where the numbers of training updates are also equal, and increasing the steps of the gradient accumulation from 1 to 2 improves the success rate by more than 14 points.  Consequently, we maintain gradient accumulation at 2 for all approaches. 

\paragraph{Interleaving in Flex.}
Unlike Yang et al.~\cite{flex}, we observe no benefit from using interleaving in Flex with FIVE-VLA (compare L14 vs. L15 and L16 vs. L17). We note several key differences in datasets and experimental setup. Critically, our training data may lack the diversity and scale necessary for the model to learn implicit vision token allocation across time steps when using interleaving (which is also a concern for training Flex without interleaving as we elaborate in the discussions below). While we train on 140 hours of Carla simulation data, Yang et al.~\cite{flex} utilise 20000 hours of real-world driving data. Additionally, Yang et al.~\cite{flex} employ open-loop evaluation metrics, while we use a closed-loop simulation benchmark. Prior work has shown weak correlation between these metric types~\cite{bench2drive}. 

\begin{figure*}[t]
\centering

\begin{minipage}{\textwidth}
    \centering
    \setlength{\tabcolsep}{0.05em}
    \captionof{table}{\textbf{Comparison with other memory mechanisms.} We replace \gls{RAM} by each in FIVE-VLA. \gls{RAM} is the smallest learned module, has the best driving performance and lowest memory time, commonly resulting in the highest throughput. mem.(ms): memory module time; mem.(\%): percentage of inference time in the memory module, $100\times\mathrm{fps}\times\mathrm{mem.(ms)}/1000$. \textbf{Bold}: Best, \underline{underlined}: Second best. No memory is included as a baseline in \textcolor{gray}{gray font}.}
    \label{tab:othermemorytypes_app}
    \scalebox{0.67}{
        \begin{tabular}{
            l
            | cc
            | ccc
            | ccc
            | ccc
            | cc
        }
            \toprule
            \multirow{2}{*}{\textbf{Memory Type}}
            & \multirow{2}{*}{\shortstack[c]{\textbf{\#Params}\\\textbf{Memory}$\downarrow$}}
            & \multirow{2}{*}{\shortstack[c]{\textbf{\#Vision}\\\textbf{Tokens}$\downarrow$}}
            & \multicolumn{3}{c|}{\textbf{T4}}
            & \multicolumn{3}{c|}{\textbf{A100}}
            & \multicolumn{3}{c|}{\textbf{H200}}
            & \multirow{2}{*}{\textbf{DS}$\uparrow$}
            & \multirow{2}{*}{\textbf{SR(\%)}$\uparrow$} \\
            & &
            & \textbf{fps}$\uparrow$ & \textbf{mem.(ms)}$\downarrow$& \textbf{mem.(\%)}$\downarrow$
            & \textbf{fps}$\uparrow$ & \textbf{mem.(ms)}$\downarrow$& \textbf{mem.(\%)}$\downarrow$
            & \textbf{fps}$\uparrow$ & \textbf{mem.(ms)}$\downarrow$& \textbf{mem.(\%)}$\downarrow$
            & & \\
            \hline
            \textcolor{gray}{No Memory}
            & \textemdash
            & \textcolor{gray}{98}
            & \textcolor{gray}{5.35} & \textemdash & \textemdash
            & \textcolor{gray}{30.85} & \textemdash & \textemdash
            & \textcolor{gray}{59.22} & \textemdash & \textemdash
            & \textcolor{gray}{88.49}
            & \textcolor{gray}{73.03} \\
            \hline
            Sliding Window
            & \textemdash
            & 392
            & 1.60 & \textemdash & \textemdash
            & 19.93 & \textemdash & \textemdash
            & 34.08 & \textemdash & \textemdash
            & 87.09
            & 69.70 \\
            \quad+efficient impl.
            &
            &
            & 3.22 & \textemdash & \textemdash
            & 28.21 & \textemdash & \textemdash
            & 56.08 & \textemdash & \textemdash
            &
            &  \\
            Flex~\cite{flex}
            & 56M
            & \underline{80}
            & 1.94 & 24.99 & 4.85
            & 19.10 & 3.06 & 5.84
            & 33.31 & 1.62 & 5.40
            & 87.14
            & 70.11 \\
            \quad+efficient impl.
            &
            &
            & 4.66 & 25.52 & 11.89
            & 26.99 & 3.06 & 8.26
            & 54.01 & 1.56 & 8.43
            &
            &  \\
            QT-Former~\cite{ORION}
            & 77M
            & \textbf{36}
            & \textbf{5.64} & 13.04 & 7.35
            & 28.01 & 4.26 & 11.93
            & 54.02 & 2.13 & 11.51
            & 87.37
            & 71.06 \\
            \hline
            \rowcolor{blue!10}
            \gls{RAM} Reset
            & \textbf{17M}
            & 98
            & \underline{5.15} & \textbf{1.43} & \textbf{0.74}
            & \textbf{30.11} & \textbf{0.56} & \textbf{1.69}
            & \textbf{58.82} & \textbf{0.32} & \textbf{1.88}
            & \underline{90.21}
            & \underline{75.91} \\
            \rowcolor{blue!10}
            \gls{RAM} Accumulation
            & \textbf{17M}
            & 2$\times$98
            & 3.89 & \underline{2.03} & \underline{0.79}
            & \underline{29.96} & \underline{0.67} & \underline{2.01}
            & \underline{56.71} & \underline{0.41} & \underline{2.33}
            & \textbf{90.95}
            & \textbf{77.27} \\
            \bottomrule
        \end{tabular}
    }
    \vspace{10pt}
\end{minipage}
\end{figure*}

\paragraph{Further discussions on efficiency.} In \cref{tab:othermemorytypes_app}, \gls{RAM} Reset takes about $18\times$ and $9\times$ less memory-module time than Flex and QT-Former on T4 (1.43\,ms vs. 25.52\,ms and 13.04\,ms, respectively). Across devices and \gls{RAM} modes, the module accounts for 0.74\%--2.33\% of inference time, but this does not measure the total overhead of recurrence. Accumulation also processes a secondary state in the \gls{LLM}. On T4, throughput is 5.15 fps with Reset, 3.89 fps with accumulation, and 5.35 fps without memory.

\paragraph{Discussions.}
The best performance achieved using vision-based memory mechanisms (DS: 86.76, SR: 71.06\% in L7) approaches but does not surpass the baseline model (L1). This underperformance can be attributed to several factors. For sliding window, we propagate 392 vision tokens to FastVLM, which was originally trained with 256 tokens (from 1024-resolution images). This exceeds the training configuration of FastVLM, which also lacks video data in its pretraining. Our training set may therefore be insufficient for the model to adapt to higher token counts or temporal sequences. For QT-Former and Flex, randomly initialised modules are inserted between the vision encoder and \gls{LLM}, disrupting the pretrained FastVLM (trained on 40M samples). Consequently, either our training dataset is inadequate for training these modules from scratch (to highlight again, Flex is trained with 20000 hours of data while we use 140 hours), or alternative training strategies, potentially involving multi-stages, are necessary. These factors require further investigation, which is beyond the scope of this work.

\subsection{Further Results on Fail2Drive and Bench2Drive Benchmarks}
\begin{table}[t]
\setlength{\tabcolsep}{2em}
\centering
\caption{Comparison with models trained longer without \gls{RAM}.}
\label{tab:longertraining}
\scalebox{0.83}{
\begin{tabular}{cc|cc} 
\toprule
\textbf{Training Length}&\textbf{\gls{RAM}}&\textbf{DS} $\uparrow$&\textbf{SR(\%)} $\uparrow$ \\ \hline
6 epochs&\xmark&88.49&73.03\\ 
10 epochs&\xmark&87.83&72.73\\
6 epochs+fine-tuning&\xmark&88.70&73.94\\ \hline 
\rowcolor{blue!10} 6 epochs+fine-tuning&\cmark&\textbf{90.95}&\textbf{77.27}\\
\bottomrule
\end{tabular}}
\end{table}

\paragraph{Comparison with longer trained-models.} We also investigate if our gain originates simply due to longer training of the model or it is the contribution of the \gls{RAM}. For this purpose, we introduce two additional baselines that are trained longer without \gls{RAM}, where we (i) train a model for 10 epochs, the same training budget with our models, and (ii) fine-tune the same checkpoint we use for the same number of training iterations we use for \gls{RAM}. \cref{tab:longertraining} shows that both options perform similar to the model trained for 6 epochs, hence FIVE-VLA did not have an unfair advantage due to further fine-tuning in the second stage.

\paragraph{Individual Ability Scores on Bench2Drive.} \cref{tab:sota_app} compares individual ability scores of FIVE-VLA with existing \gls{E2E}-\gls{AD} and \glspl{VLA}, again showing the superiority of FIVE-VLA.

\begin{table}[t]
\centering
\caption{\textbf{Comparison with \gls{SOTA}} on Bench2Drive in terms of Driving Score (DS), Success Rate (SR\%), and specialised abilities. FIVE-VLA reaches \gls{SOTA}, outperforming all approaches. M: Multi-view, S: Single view, L: LiDAR. 
Improvements (in \textcolor{forestgreen}{green} and \textcolor{red}{red}) are specified in comparison to the \underline{underlined} best camera-only baseline.}
\label{tab:sota_app}
\scalebox{0.59}{
\begin{tabular}{lcc|cc|ccccc|c|c}
\toprule
& & & & & \multicolumn{6}{c|}{\textbf{Ability}$\uparrow$} \\
\cline{6-11}
\textbf{Method} & \textbf{Sensors} & \textbf{CoT} & \textbf{DS} $\uparrow$ & \textbf{SR(\%)} $\uparrow$ & \textbf{Merging} & \makecell{\textbf{Over-}\\\textbf{taking}} & \makecell{\textbf{Emergency}\\\textbf{Brake}} & \makecell{\textbf{Give}\\\textbf{Way}} & \makecell{\textbf{Traffic}\\\textbf{Sign}} & \textbf{Mean} & \textbf{Venue} \\
\hline
\multicolumn{12}{l}{\textbf{E2E-AD Approaches}} \\
VAD~\cite{vad} & M & \textemdash & 42.35 & 15.00 & 8.11 & 24.44 & 18.44 & 20.00 & 19.15 & 18.07 & ICCV23 \\
DriveTransformer~\cite{drivetransformer} & M & 	\textemdash & 63.46 & 35.01 & 17.57 & 35.00 & 48.36 & 40.00 & 52.10 & 38.60 & ICLR25 \\
Raw2Drive~\cite{raw2drive} & M & 	\textemdash & 71.36 & 50.24 & 43.35 & 51.11 & 60.00 & 50.00 & 62.26 & 53.34 & NeurIPS25 \\
VADv2~\cite{vadv2} & M & 	\textemdash & 76.15 & 50.46 & \textemdash & \textemdash & \textemdash & \textemdash & \textemdash & \textemdash & ICLR26 \\
PGS~\cite{PGS} & M & \textemdash & 78.08 & 48.64 & 35.00 & \underline{73.33} & 55.00 & \underline{60.00} & 43.68 & 53.40 & NeurIPS25 \\
GaussianFusion~\cite{gaussianfusion} & S, L & \textemdash & 79.10 & 54.40 & 36.60 & 64.40 & 66.50 & 53.30 & 60.80 & 56.30 & NeurIPS25 \\
Transfuser++~\cite{transfuserplusplus} & S, L & \textemdash & 84.21 & 67.27 & 58.75 & 57.77 & 83.33 & 40.00 & 82.11 & 64.39 & ICCV23 \\
BridgeDrive~\cite{bridgedrive} & S, L & \textemdash & 87.99 & 74.99 & \textbf{69.92} & 66.67 & \textbf{90.00} & 50.00 & \textbf{89.47} & 73.15 & ICLR26 \\
\hline
\multicolumn{12}{l}{\textbf{Vision-Language-Action Models}} \\
DriveMoE~\cite{drivemoe} & M & \xmark & 74.22 & 48.64 & 34.67 & 40.00 & 65.45 & 40.00 & 59.44 & 47.91 &CVPR26\\
ORION~\cite{ORION} & M & \xmark & 77.74 & 54.62 & 25.00 & 71.11 & 78.33 & 30.00 & 69.15 & 54.72 & ICCV25\\
AutoVLA~\cite{autovla} & M & \cmark & 78.84 & 57.73 & \textemdash & \textemdash & \textemdash & \textemdash & \textemdash & \textemdash &NeurIPS25\\
SimLingo~\cite{SimLingo} & S & \cmark & \underline{85.07} & \underline{67.27} & \underline{54.01} & 57.04 & \underline{88.33} & 53.33 & \underline{82.45} & \underline{67.03} &CVPR25\\
\hline
\rowcolor{blue!10}
FIVE-VLA&S&\xmark&\textbf{90.95$\pm$1.2}&\textbf{77.27$\pm$1.6}&60.00$\pm$3.3&\textbf{83.70$\pm$3.4}&87.78$\pm$2.6&\textbf{76.67$\pm$5.8}&85.61$\pm$1.9&\textbf{78.75$\pm$0.4}&Ours\\
\rowcolor{blue!10}
&&&\imp{5.88}&\imp{10.00}&\imp{5.99}&\imp{10.37}&\nimp{0.55}&\imp{16.67}&\imp{3.16}&\imp{11.72}&\\
\bottomrule
\end{tabular}
}
\end{table}

\begin{figure*}[t]
\centering

\begin{minipage}{\textwidth}
\centering
\captionof{table}{\textbf{Results on Fail2Drive.} In-Distribution evaluates on known CARLA scenarios; Generalisation measures robustness in \gls{OOD} scenarios. Visual-lon/lat denote longitudinal/lateral perception scenarios. M: Multi-view, S: Single view, L: LiDAR. Improvements (in \textcolor{forestgreen}{green} and \textcolor{red}{red}) are specified in comparison to the \underline{underlined} best camera-only baseline.}
\label{tab:fail2drive_app}
\setlength{\tabcolsep}{0.1em}
\scalebox{0.74}{
\begin{tabular}{l|c|ccc|ccc|cccc}
\toprule
\multirow{2}{*}{\textbf{Method}}&\multirow{2}{*}{\textbf{Sensors}}& \multicolumn{3}{c|}{\textbf{In-Distribution}} & \multicolumn{3}{c|}{\textbf{Generalisation}}& \multicolumn{4}{c}{\textbf{Per-class metrics (\textbf{HM} $\uparrow$ )}} \\
\cmidrule{3-12}
& & \textbf{DS} $\uparrow$ & \textbf{SR(\%)} $\uparrow$ & \textbf{HM} $\uparrow$ & \textbf{DS} $\uparrow$ & \textbf{SR(\%)} $\uparrow$ & \textbf{HM} $\uparrow$ & \textbf{Visual-lon} & \textbf{Visual-lat} & \textbf{Behaviour} & \textbf{Robustness} \\
\midrule
Transfuser++ & S, L & 83.3 & 78.5 & 80.8 & 75.4 \scriptsize{(-9.5\%)} & 61.1 \scriptsize{(-22.2\%)} & 67.5 \scriptsize{(-16.5\%)} & \textbf{77.0} \scriptsize{(-7.8\%)} & 40.4 \scriptsize{(-30.6\%)} & 47.2 \scriptsize{(-43.9\%)} & 93.7 \scriptsize{(-2.2\%)} \\
\midrule
TCP & S & 24.7 & 39.1 & 30.3 & 24.5 \scriptsize{(-0.8\%)} & 31.4 \scriptsize{(-19.7\%)} & 27.5 \scriptsize{(-9.1\%)} & 31.4 \scriptsize{(-10.3\%)} & 6.2 \scriptsize{(-1.4\%)} & 22.1 \scriptsize{(-30.6\%)} & 42.8 \scriptsize{(3.6\%)} \\
UniAD & M & 47.5 & 36.3 & 41.2 & 44.0 \scriptsize{(-7.4\%)} & 27.6 \scriptsize{(-24.0\%)} & 33.9 \scriptsize{(-17.6\%)} & 26.3 \scriptsize{(4.6\%)} & 13.0 \scriptsize{(84.2\%)} & 17.9 \scriptsize{(-67.9\%)} & 66.3 \scriptsize{(-4.7\%)} \\
ORION & M & 53.0 & 52.0 & 52.5 & 51.2 \scriptsize{(-3.4\%)} & 46.0 \scriptsize{(-11.5\%)} & 48.5 \scriptsize{(-7.7\%)} & 53.0 \scriptsize{(-9.6\%)} & 34.0 \scriptsize{(47.9\%)} & 35.4 \scriptsize{(-34.0\%)} & 66.0 \scriptsize{(-8.4\%)} \\
HiP-AD & M & 74.1 & 70.7 & 72.4 & 67.1 \scriptsize{(-9.4\%)} & \underline{56.7} \scriptsize{(-19.8\%)} & 61.5 \scriptsize{(-15.1\%)} & 70.9 \scriptsize{(-5.6\%)} & 40.8 \scriptsize{(-27.1\%)} & \underline{42.6} \scriptsize{(-46.6\%)} & 82.3 \scriptsize{(4.1\%)} \\
SimLingo & S & \underline{82.6} & \underline{79.3} & \underline{80.9} & \underline{71.7} \scriptsize{(-13.2\%)} & 55.0 \scriptsize{(-30.6\%)} & \underline{62.2} \scriptsize{(-23.1\%)} & \underline{71.1} \scriptsize{(-9.0\%)} & \underline{45.9} \scriptsize{(-32.2\%)} & 31.2 \scriptsize{(-64.2\%)} & \underline{86.8} \scriptsize{(-5.9\%)} \\
\midrule
\rowcolor{blue!10}
FIVE-VLA & S & \textbf{86.0} & \textbf{83.7} & \textbf{84.8} & \textbf{77.5} \scriptsize{(-9.9\%)} & \textbf{67.0} \scriptsize{(-20.0\%)} & \textbf{71.9} \scriptsize{(-15.3\%)} & 73.9 \scriptsize{(-9.3\%)} & \textbf{60.6} \scriptsize{(-16.5\%)} &  \textbf{51.3} \scriptsize{(-43.2\%)} &  \textbf{94.6} \scriptsize{(-1.8\%)} \\
\rowcolor{blue!10}
& & \imp{3.4} & \imp{4.4} & \imp{3.9} & \imp{5.8} & \imp{10.3} & \imp{9.7} & \imp{2.8} & \imp{14.7} & \imp{8.7} & \imp{7.8} \\
\bottomrule
\end{tabular}
}
\end{minipage}

\begin{minipage}{\textwidth}
\centering
\includegraphics[width=0.98\textwidth]{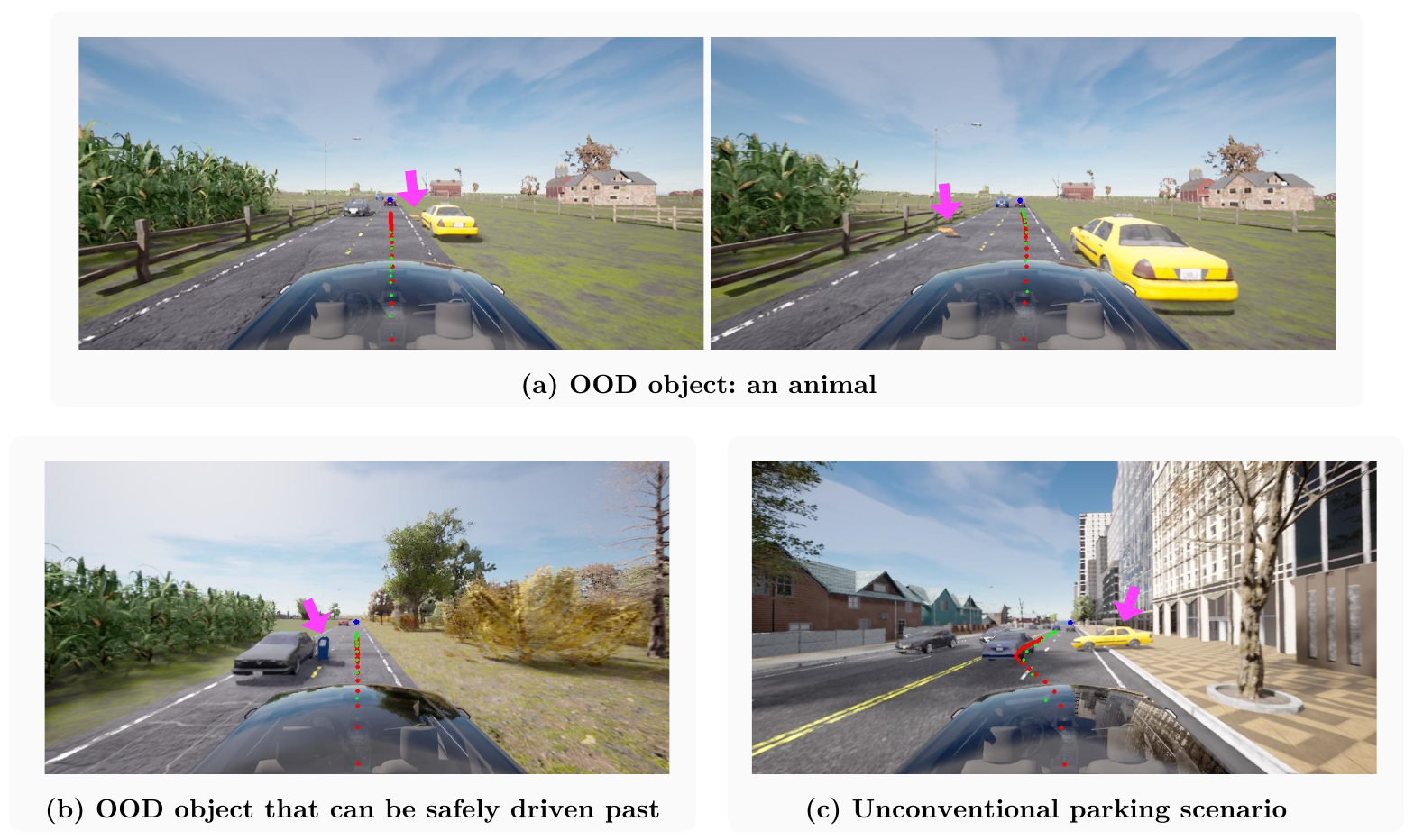}
\captionof{figure}{\textbf{Qualitative examples on Fail2Drive scenarios.} \textbf{(a)}~The ego decelerates as an animal crosses the road and resumes once the path is clear; an unusually positioned vehicle is also parked on the roadside. These animals and parking cars on the roadside do not exist in the training set, making this an \gls{OOD} scenario. \textbf{(b)}~A bin as an \gls{OOD} object lies outside the drivable corridor, so the ego proceeds without deviation. \textbf{(c)}~The ego correctly plans a lane change to avoid a vehicle parked in an unconventional manner, unseen during training. Objects of interest are indicated by a \textcolor{magenta}{magenta} arrow; \textcolor{red}{red}: path waypoints, \textcolor{forestgreen}{green}: speed waypoints, \textcolor{blue}{blue}: target points.}
\label{fig:fail2drive_examples}
\end{minipage}

\end{figure*}

\paragraph{Detailed Fail2Drive Results.} Beyond the aggregate scores presented in \cref{sec:experiments,tab:sota}, \cref{tab:fail2drive_app} decomposes generalisation performance along Fail2Drive's four evaluation axes: \textit{Visual-Longitudinal}, \textit{Visual-Lateral}, \textit{Behaviour}, and \textit{Robustness}. FIVE-VLA improves over the best camera-only baseline on every axis, with the largest gains on \textit{Visual-Lateral} (60.6 vs.\ 45.9, +14.7), \textit{Behaviour} (51.3 vs.\ 42.6, +8.7), and \textit{Robustness} (94.6 vs.\ 86.8, +7.8); the last drops only 1.8\% relative to in-distribution, suggesting that FIVE-VLA does not overreact to benign novelty. \cref{fig:fail2drive_examples} illustrates these strengths qualitatively across three of the four axes: in (a), the ego decelerates due to a very small animal crossing the road, a class absent from training, and resumes once the path is clear, a \textit{Visual-Longitudinal} case; in (b), it correctly ignores an irrelevant \gls{OOD} object that does not obstruct its lane, the kind of distractor probed by the \textit{Robustness} axis; and in (c), it executes a lane change around a vehicle parked at an unconventional angle, the ``BadParking'' case targeted by the \textit{Visual-Lateral} axis.

\subsection{Further Discussions on Efficiency}
\begin{figure*}[t]
\centering
\begin{minipage}{\linewidth}
\centering
\captionof{table}{\textbf{Effect of text generation on throughput for SimLingo and FIVE-VLA on a T4 GPU.} We report fps and, in parentheses, the mean number of sequential \gls{LLM} forward passes per trajectory. \textit{Efficient Driving} supplies the fixed four-token prefix as input, requiring one pass; \textit{Driving} generates the four tokens and then predicts the trajectory, requiring five (\cref{subsec:capabilities}). \textcolor{gray}{Gray font} shows modes not available in the standard implementation of a model. Text generation substantially reduces the throughput. FIVE-VLA achieves $\sim 3\times$ higher throughput than SimLingo across all modes. \gls{RAM} accumulation is used with FIVE-VLA.}
\setlength{\tabcolsep}{1.8em}
\scalebox{0.9}{
\begin{tabular}{lccc}
\toprule
\textbf{Model} & \textbf{Efficient Driving} & \textbf{Driving} & \textbf{Driving with CoT} \\
\hline
SimLingo & \textcolor{gray}{1.06 (1)}& 0.50 (5)& 0.13 (25.30)\\
\rowcolor{blue!10} FIVE-VLA & \textbf{3.89} (1)& \textcolor{gray}{\textbf{1.46} (5)}& \textbf{0.39} (24.55)\\
\bottomrule
\end{tabular}
}
\label{tab:throughput}
\end{minipage}
\end{figure*}

\paragraph{The Impact of Autoregressive Text Generation on Efficiency}
To quantify the effect of the \gls{AR} text generation for FIVE-VLA and SimLingo, we measure throughput on a T4 GPU across three operating modes: efficient driving, driving, and driving with \gls{CoT}. \textit{Efficient driving} obtains trajectories in one \gls{LLM} forward pass without text generation. \textit{Driving} generates the four-token ``Waypoints:''/EOS prefix in four passes and processes the action queries in a fifth; \textit{driving with \gls{CoT}} adds a variable number of reasoning tokens. When both \gls{RAM} accumulation states are active, they are batched in the trajectory-prediction call, not counted as two sequential passes.

To demonstrate the effect clearly also by comparing FIVE-VLA with SimLingo, we modify each model to support all three modes. For SimLingo's efficient driving mode, we append four tokens to the input text  (\texttt{``Way''}, \texttt{``points''}, \texttt{``:''}, \texttt{<EOS>}) during inference following \cref{subsec:capabilities}. Conversely, for FIVE-VLA's driving mode, we remove these tokens and follow SimLingo's standard implementation.

\cref{tab:throughput} presents the results, demonstrating that text generation substantially reduces throughput. Compared to efficient driving, the superfluous text output in driving mode decreases throughput by more than $2\times$ for both models (e.g., from 3.89 fps to 1.46 fps for FIVE-VLA). The impact is even more pronounced with \gls{CoT}, where each model requires approximately 25 passes to generate a trajectory, resulting in nearly $10\times$ lower throughput (from 3.89 fps to 0.39 fps for FIVE-VLA). This highlights a critical challenge for deploying \gls{CoT} reasoning in real-world applications, despite its benefits for explainability. Finally, FIVE-VLA maintains approximately $3\times$ higher throughput than SimLingo across all modes, demonstrating the efficiency gains attributable to FIVE-VLA's architectural design.

\begin{figure*}[t]
\centering
\begin{minipage}{\textwidth}
    \centering
    \captionof{table}{\textbf{Space and time efficiency comparison with \glspl{VLA}.} \#Tokens includes the number of tokens propagated to the \gls{LLM}. For SimLingo and ORION, we measure time efficiency and \#Tokens without \gls{CoT} for a fair comparison. Mem.(GB) denotes peak GPU VRAM utilisation on A100 during inference. OOM: Out-of-memory for ORION on T4. $^*$: As trained models are not public for AutoVLA and DriveMoE, the throughput and \#Tokens are incomplete. AutoVLA reports $\sim 1$ fps as throughput, although the GPU type is not specified. DriveMoE does not report throughput.
     }
    \setlength{\tabcolsep}{0.5em}
    \label{tab:efficiency_app}
    \scalebox{0.9}{
    \begin{tabular}{l|ccc|ccc|cc}
    \toprule
    \multirow{2}{*}{\textbf{Method}}&\multicolumn{3}{c|}{\textbf{Space}}&\multicolumn{3}{c|}{\textbf{Throughput in fps $\uparrow$}}&\multirow{2}{*}{\textbf{DS}$\uparrow$}&\multirow{2}{*}{\textbf{SR(\%)} $\uparrow$} \\
    &\textbf{\#Params} $\downarrow$&\textbf{\#Tokens} $\downarrow$& \textbf{VRAM (GB)} $\downarrow$&\textbf{T4}&\textbf{A100}&\textbf{H200}&& \\ \hline 
    DriveMoE$^*$~\cite{drivemoe}&3B& \textemdash & \textemdash & \textemdash &\textemdash&\textemdash&74.22&48.64\\
    ORION~\cite{ORION}&7B& 599 & 28.93 & OOM & 0.98 & 2.13&77.74&54.62\\
    AutoVLA$^*$~\cite{autovla}&3B & \textemdash &\textemdash&\textemdash&$\sim$1&\textemdash&78.84&57.73\\
    SimLingo~\cite{SimLingo}&1B&577 & 2.02 &0.50 & 7.99 &  14.14 &85.88&68.18\\
    \hline
    \rowcolor{blue!10} 
    FIVE-VLA &\textbf{641M}& \textbf{2x164} & \textbf{1.68} & \textbf{3.89} & \textbf{29.96} & \textbf{56.71} &\textbf{90.95}&\textbf{77.27} \\
    \bottomrule
    \end{tabular}}
\end{minipage}

\end{figure*}

\paragraph{Full version of  \cref{tab:efficiency}.} \cref{tab:efficiency_app} provides a full version of  \cref{tab:efficiency} excluded from the main text considering the space constraints. Additionally, this includes the peak VRAM and the total number of tokens propagated through the \gls{LLM}. In the single-view closed-loop configuration, FIVE-VLA forwards $2\times164$ tokens, compared to 577 for SimLingo and 599 for ORION. It operates within a 1.68GB VRAM budget, compared to 2.02GB for SimLingo, and 28.93GB for ORION which exceeds the 16GB VRAM capacity of a Tesla T4 and consistent with the OOM observed at the edge tier.

\subsection{Language Capabilities}
Here, we include qualitative and quantitative results to demonstrate the language capabilities of the model trained for closed-loop driving. To this end, we report quantitative \gls{VQA} results, present closed-loop driving results with \gls{CoT} reasoning, include qualitative results for \gls{VQA} and using high-level commands for navigational conditioning.

\subsubsection{DriveLM-CARLA VQA Evaluation}
\label{app:drivelmvqa}
\textbf{Dataset and inference.} We evaluate FIVE-VLA and SimLingo on DriveLM-CARLA~\cite{drivelm}, without additional training or adaptation for this evaluation. We use the same 200 randomly selected routes for both models, comprising 9,428 key frames and 266,952 question--answer pairs spanning perception (46\%), prediction (32\%), and planning (22\%). Each question is answered independently using the front-camera image and recorded ego-state inputs through the model's \gls{VQA} interface, with memory state reset between questions. This assesses the existing checkpoints' driving-related language capability and it is not an evaluation of temporal \gls{RAM} reasoning in the answer stream.

\textbf{Scoring protocol.} Following DriveLM's use of complementary language metrics, we report SPICE, BLEU-4, ROUGE-L, CIDEr, and METEOR, together with an LLM Judge-score. We replace the ChatGPT judge with the open-weight Qwen3-30B-A3B-Instruct-2507 model~\cite{qwen3}\footnote{\url{https://huggingface.co/Qwen/Qwen3-30B-A3B-Instruct-2507}}, applying the same reference-answer rubric to both models. The judge receives the generated and ground-truth answers and scores content agreement from 0 to 100, explicitly ignoring sentence structure and assigning zero to completely different content; it does not inspect the image or independently verify the driving scene. The scoring implementation uses temperature 0.6 with thinking disabled and averages valid returned scores. Standard captioning metrics use the same answer normalisation and evaluator for both models: answer prefixes and end-of-message markers are removed, and the evaluator processes chunks of up to 500 pairs, combining chunk-level metrics weighted by the number of pairs. Only Judge-score is reported on a 0-100 scale; the remaining metrics retain their evaluator scales.

\begin{table}[t]
\centering
\small
\setlength{\tabcolsep}{0.6em}
\caption{\textbf{Language performance on DriveLM-CARLA without additional training.} Both models answer the same 266,952 VQA pairs. All metrics are higher-is-better; Judge-score uses Qwen3 rather than ChatGPT and is on a 0--100 scale. The last row gives absolute differences computed from the displayed scores, not relative percentages.}
\label{tab:drivelm_vqa}
\resizebox{\linewidth}{!}{%
\begin{tabular}{lcccccc}
\toprule
\textbf{Model} & \textbf{Judge-score} & \textbf{SPICE} & \textbf{BLEU-4} & \textbf{ROUGE-L} & \textbf{CIDEr} & \textbf{METEOR} \\
\midrule
SimLingo~\cite{SimLingo} & 71.80 & 0.8724 & 0.7207 & 0.8477 & 7.0825 & 0.5057 \\
\rowcolor{blue!10}
FIVE-VLA (Ours) & \textbf{82.76} & \textbf{0.9113} & \textbf{0.8164} & \textbf{0.9023} & \textbf{8.0742} & \textbf{0.5754} \\
\midrule
Absolute gain & +10.96 & +0.0389 & +0.0957 & +0.0546 & +0.9917 & +0.0697 \\
\bottomrule
\end{tabular}}
\end{table}

\textbf{Results and scope.} FIVE-VLA outperforms SimLingo on all six reported metrics in \cref{tab:drivelm_vqa}, including a 10.96-point gain in Judge-score. The complementary gains in semantic-content metrics and lexical-overlap metrics support stronger agreement with reference answers, providing evidence that efficient trajectory prediction need not come at the expense of driving-related language capability in this evaluation.

\begin{table}[t]
\centering
\caption{Closed-loop evaluation results on Bench2Drive benchmark with different driving modes of SimLingo and FIVE-VLA.}
\setlength{\tabcolsep}{1.3em}
\label{tab:ablation}
\scalebox{0.9}{
\begin{tabular}{cccc|cc} 
\midrule
\textbf{Model}&\textbf{Driving Mode}&\textbf{FastVLM}&\textbf{\gls{RAM}}&\textbf{DS} $\uparrow$ & \textbf{SR(\%)} $\uparrow$ \\ \midrule
\multirow{2}{*}{SimLingo} &Driving w.\gls{CoT}&\xmark&\xmark&85.84&66.36\\
&Driving&\xmark&\xmark&85.88&68.18\\ \hline 
\multirow{3}{*}{FIVE-VLA}&Efficient Driving&\cmark&\xmark&88.49&73.03\\ 
&Driving w.\gls{CoT}&\cmark&\cmark&88.74&73.94\\
&Efficient Driving&\cmark&\cmark&\textbf{90.95}&\textbf{77.27}\\
\midrule
\end{tabular}
}
\end{table}

\subsubsection{Closed-Loop Driving with \gls{CoT} Reasoning.} \cref{tab:ablation} compares how driving performance of SimLingo and FIVE-VLA changes when driving with \gls{CoT} is used. We observe that the driving performance of the models slightly degrades for both models. For example, the success rate drops around 1.8 points for SimLingo and 3.3 points for FIVE-VLA when driving with \gls{CoT} reasoning is used. However, the gap between SimLingo and FIVE-VLA still remains substantial. 

\subsubsection{Qualitative Examples}
In this section, we provide qualitative results obtained with our model and compare them with SimLingo. These experiments suggest that FIVE-VLA preserves the language capabilities of SimLingo in \gls{CoT} reasoning, \gls{VQA} and the ability to process high-level commands.

\begin{figure*}[htpb]
\begin{minipage}{\textwidth}
        \centering
        \includegraphics[width=0.98\textwidth]{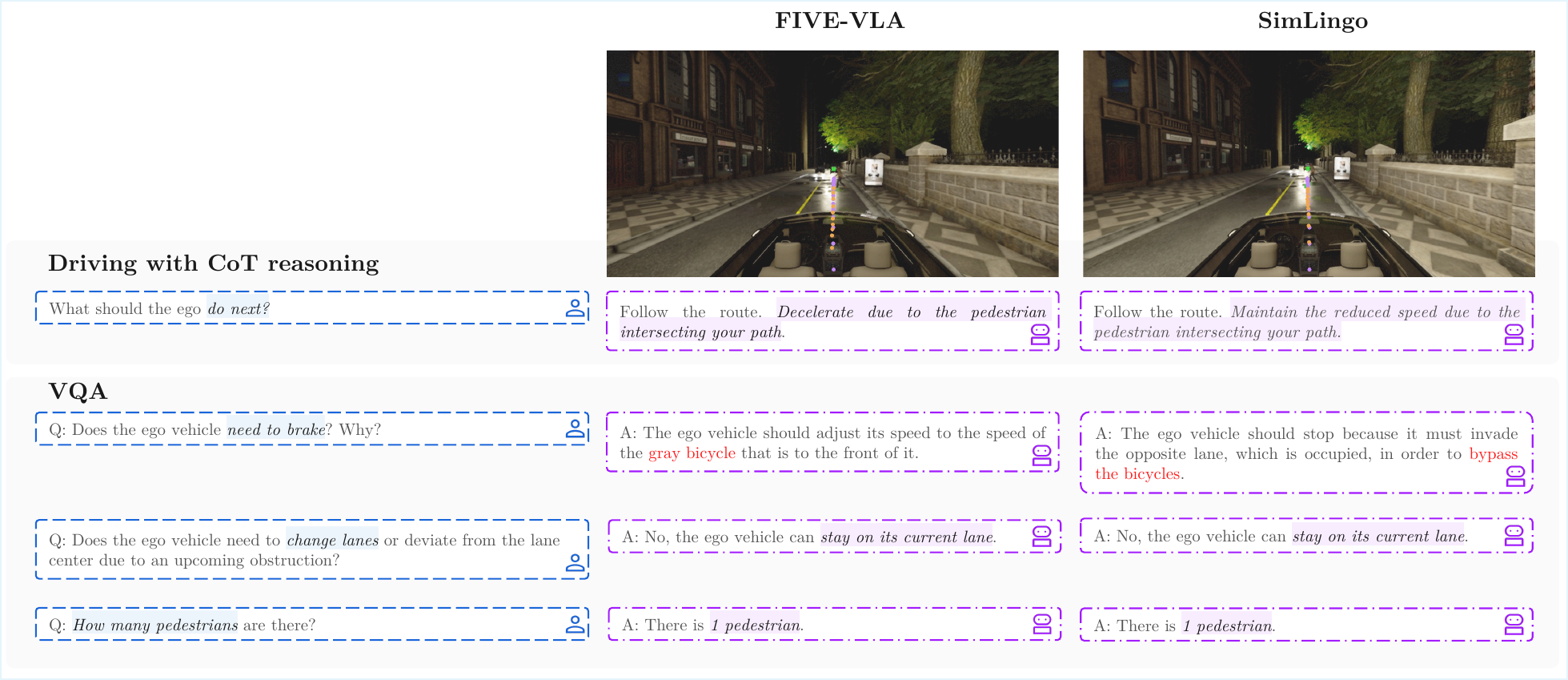}
        \captionof{figure}{\textbf{Qualitative example for \gls{CoT} reasoning and \gls{VQA}.} A night scene where a pedestrian crossing the street. The trajectory is obtained in \gls{CoT} reasoning mode. \textcolor{lightpurple}{purple}: path waypoints, \textcolor{orange}{orange}: speed waypoints, \textcolor{green}{green}: target points.}
        \label{fig:supp:comm_qa_night_pedestrian}
\end{minipage}

\begin{minipage}{\textwidth}
        \centering
        \includegraphics[width=0.98\textwidth]{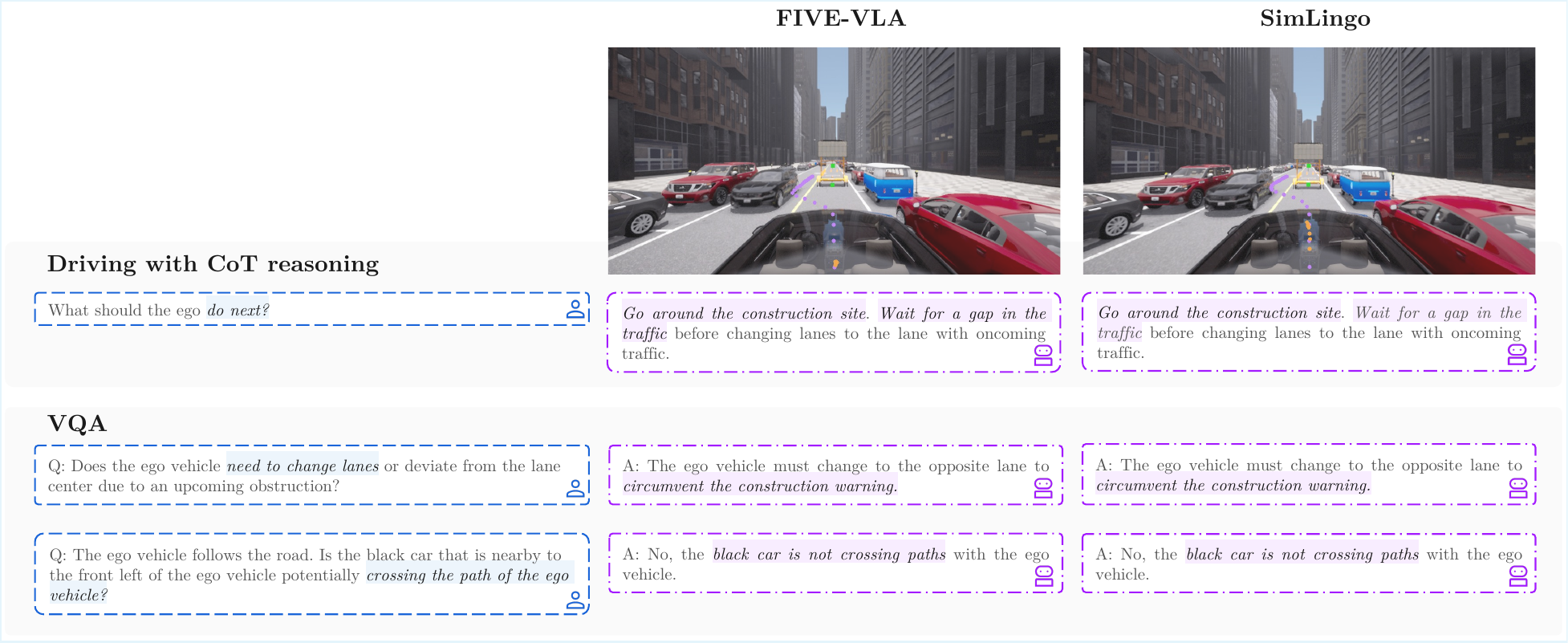}
        \captionof{figure}{\textbf{Qualitative example for \gls{CoT} reasoning and \gls{VQA}.} An urban driving scene with construction zone blocking the road. The trajectory is obtained in \gls{CoT} reasoning mode. \textcolor{lightpurple}{purple}: path waypoints, \textcolor{orange}{orange}: speed waypoints, \textcolor{green}{green}: target points.}        \label{fig:supp:comm_qa_construction}
\end{minipage}

\begin{minipage}{\textwidth}
        \centering
        \includegraphics[width=0.98\textwidth]{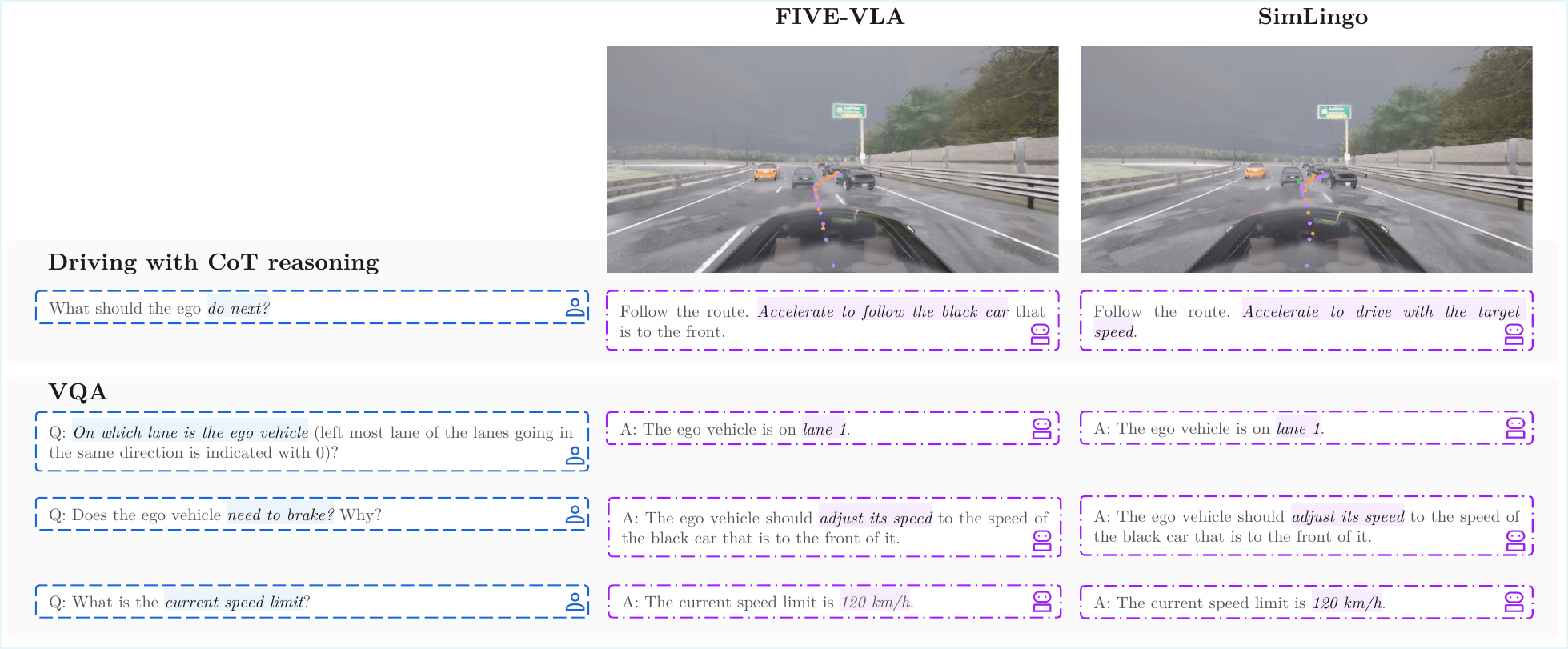}
        \captionof{figure}{\textbf{Qualitative example for \gls{CoT} reasoning and \gls{VQA}.} A scene where the ego should exit the highway. The trajectory is obtained in \gls{CoT} reasoning mode. \textcolor{lightpurple}{purple}: path waypoints, \textcolor{orange}{orange}: speed waypoints, \textcolor{green}{green}: target points.}        \label{fig:supp:comm_qa_highwayexit}
\end{minipage}
\end{figure*}

\paragraph{\gls{CoT} Reasoning and \gls{VQA}.}
As qualitative examples, we provide \gls{CoT} reasoning also by including a total of 8 \gls{VQA} examples across three different scenarios. We collect these scenarios randomly while paying attention to their diversity (night time pedestrian crossing, urban driving with construction zone blocking the road, highway exit). We compare FIVE-VLA with SimLingo in all cases to provide relative insights on the performance of our model. Overall, we observe that both models produce very similar textual responses. We also notice that the trajectories generated in \gls{CoT} reasoning mode and \gls{VQA} are very similar, thus we only display the former in \cref{fig:supp:comm_qa_night_pedestrian}-\cref{fig:supp:comm_qa_highwayexit}, for brevity.

In \cref{fig:supp:comm_qa_night_pedestrian}, we present a nighttime driving scenario where a pedestrian is crossing the route. In the commentary section, when asked ``What should the ego do next?'', FIVE-VLA provides a more comprehensive response (``Follow the route. \textit{Decelerate due to the pedestrian intersecting your path.}'') compared to SimLingo (``Follow the route. \textit{Maintain the reduced speed due to the pedestrian intersecting your path.}''). FIVE-VLA correctly identifies the need to decelerate, while SimLingo suggests maintaining an already reduced speed, which may not adequately respond to the dynamic situation. For the \gls{VQA} questions, both models demonstrate similar perception capabilities. When asked ``Does the ego vehicle need to brake? Why?'', both models hallucinate a gray bicycle in the scene. Regarding lane changes, FIVE-VLA correctly states the ego vehicle can ``stay on its current lane,'' matching SimLingo's response. Both models accurately identify ``1 pedestrian'' when asked ``How many pedestrians are there?'', demonstrating consistent perception accuracy. Overall, in this example, FIVE-VLA shows more appropriate situational reasoning, particularly in identifying the immediate need to decelerate for the pedestrian.

Second, in \cref{fig:supp:comm_qa_construction}, we present an urban driving scenario with a construction zone blocking the road, requiring the ego to navigate around the obstruction. In driving with \gls{CoT} reasoning, both models provide identical and appropriate responses: ``Go around the construction site. Wait for a gap in the traffic before changing lanes to the lane with oncoming traffic.'' This shows that both models correctly understand the need to wait for safe traffic conditions before executing the lane change manoeuvre around the construction zone. For the \gls{VQA}, both models show consistent reasoning capabilities as well. When asked ``Does the ego vehicle need to change lanes or deviate from the lane centre due to an upcoming construction?'', both models correctly respond that ``The ego vehicle must change to the opposite lane to circumvent the construction warning,'' demonstrating accurate scene understanding and appropriate action planning. Regarding the potential threat from nearby vehicles, when asked ``The ego vehicle follows the road. Is the black car that is nearby to the front left of the ego vehicle potentially crossing the path of the ego vehicle?'', both FIVE-VLA and SimLingo correctly identify that "No, the black car is not crossing paths with the ego vehicle." This suggests that both models are able to handle spatial relationships and provide safe and contextually appropriate responses in this scenario.

Finally, in \cref{fig:supp:comm_qa_highwayexit}, we present a highway scenario where the ego vehicle should exit the highway. As the \gls{CoT} reasoning, FIVE-VLA responds ``Follow the route. Accelerate to follow the black car that is to the front,'' while SimLingo provides ``Follow the route. Accelerate to drive with the target speed.'' Both responses appropriately identify the need to accelerate; however, FIVE-VLA's reasoning is more specific by referencing the black car ahead, while SimLingo focuses on matching the target speed, which may be more aligned with general highway driving principles. For the \gls{VQA}, both models demonstrate consistent perception and reasoning. When asked ``On which lane is the ego vehicle (leftmost lane of the lanes going in the same direction is indicated with 0)?'', both correctly identify ``The ego vehicle is on lane 1.'' Regarding braking, when asked ``Does the ego vehicle need to brake? Why?'', both models provide identical responses: ``The ego vehicle should adjust its speed to the speed of the black car that is to the front of it,'' showing appropriate behaviour. For the speed limit question ``What is the current speed limit?'', both models accurately respond ``The current speed limit is 120 km/h,'' reflecting appropriate use of contextual cues. Overall, FIVE-VLA shows comparable performance to SimLingo in this highway scenario, with both models providing safe driving suggestions grounded in the environmental context.

\begin{figure*}[t]
        \centering
        \includegraphics[width=0.98\textwidth]{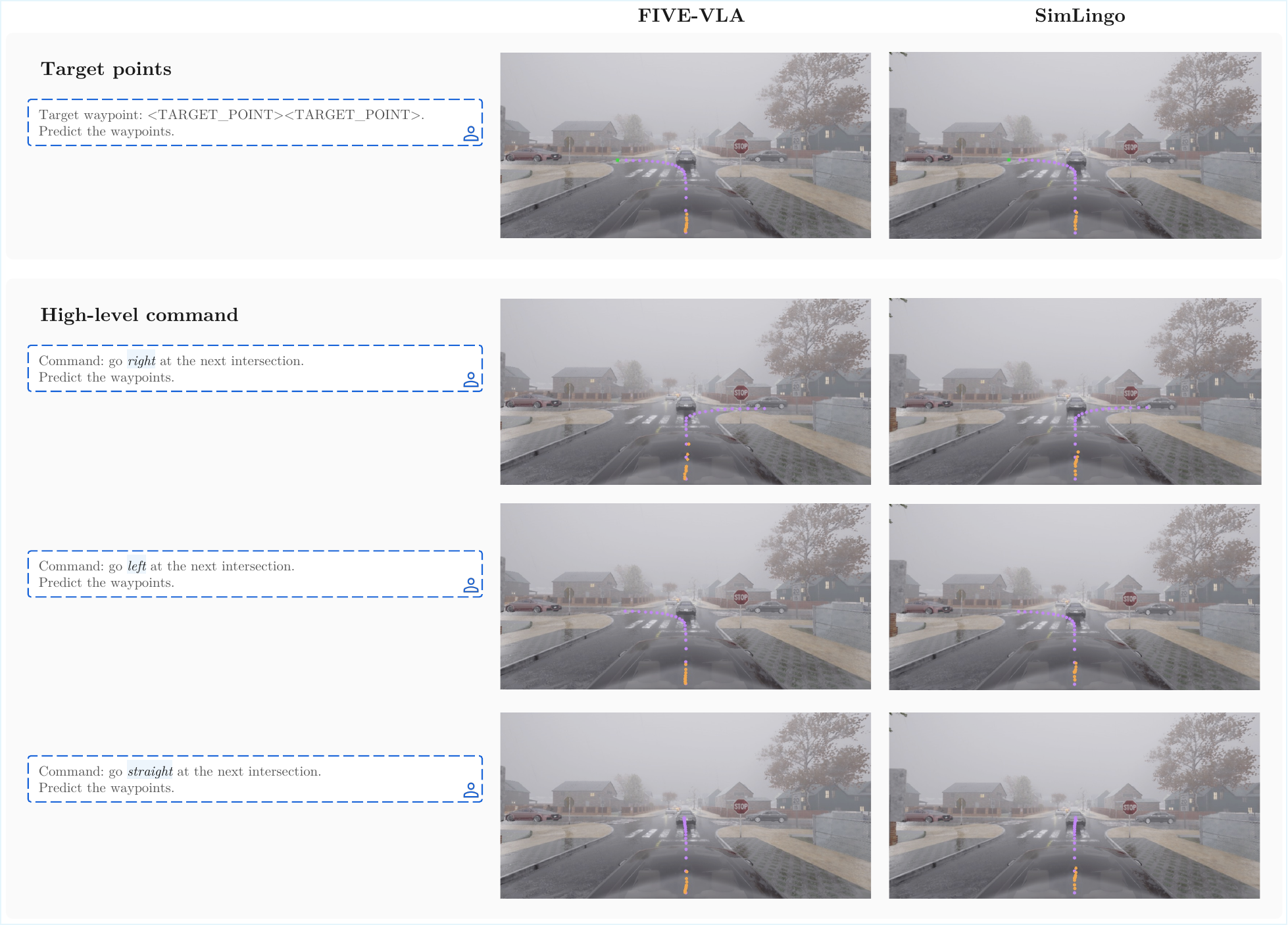}
        \vspace{-5pt}
        \caption{\textbf{Qualitative example for following high-level commands in an intersection.} The path waypoints are planned accordingly, while the speed waypoints keep the vehicle stopped considering the stop sign ahead. \textcolor{lightpurple}{purple}: path waypoints, \textcolor{orange}{orange}: speed waypoints, \textcolor{green}{green}: target points.}
        \label{fig:supp:hlc_intersection}
\end{figure*}

\begin{figure*}[t]
        \centering
        \includegraphics[width=0.98\textwidth]{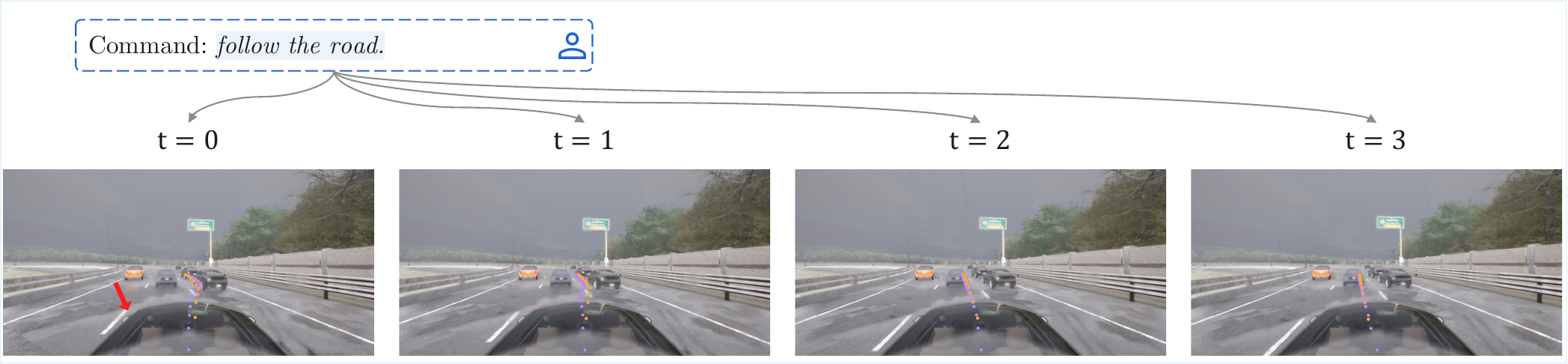}
        \vspace{-5pt}
        \caption{\textbf{Qualitative example showing the smooth change in trajectory across time steps thanks to \gls{RAM}} The model is prompted to follow the road, instead of being guided to exit the highway. For the highway exit, the ego vehicle has begun changing lanes to the right, as evidenced by the lane gap highlighted with a red arrow at $t=0$. When prompted to ``follow the road'', FIVE-VLA predicts a trajectory that \emph{smoothly} adapts over time.}
        \label{fig:supp:smoothness}
\end{figure*}

\paragraph{High-level Commands.} We present two qualitative examples to demonstrate the capability of FIVE-VLA in following high-level commands for navigational conditioning instead of target points. In \cref{fig:supp:hlc_intersection}, we show an intersection scenario where FIVE-VLA conditions on target points as well as different high-level commands. In the first row with target points (in green), the models generate similar trajectories (i.e., path waypoints). When switching to high-level command conditioning, FIVE-VLA demonstrates its ability to adapt its trajectory based on different navigational instructions. In the second row, when given the command ``go right at the next intersection,'' FIVE-VLA generates a path that appropriately curves to the right. In the third row, with the command ``go left at the next intersection,'' FIVE-VLA produces a trajectory that curves to the left, with the path waypoints clearly deviating leftward from the straight trajectory. Finally, in the fourth row, when instructed to ``go straight at the next intersection,'' FIVE-VLA maintains a path similar to the baseline, continuing straight through the intersection. Please note that the speed waypoints prevent the ego from moving considering the stop sign regardless of the navigational command. 

In \cref{fig:supp:smoothness}, we demonstrate FIVE-VLA's smooth adaptation of its trajectory when following high-level commands thanks to \gls{RAM}. Specifically, when given the high-level command "follow the road," FIVE-VLA gradually overrides the initial intent which was guided by previous target points, with the trajectory following the road at $t=3$.

These examples demonstrate that FIVE-VLA successfully interprets and executes high-level navigational commands adjusting its planned trajectory accordingly while maintaining safe driving behaviour. The model shows consistent understanding of directional commands and translates them into appropriate path planning decisions at the intersection.

\subsection{Main Limitations}
\label{subsec:limitations}
Here we discuss two main limitations of the FIVE-VLA.

\textbf{Short temporal memory.} RAM conditions action prediction on action tokens from the preceding time step, aggregating a temporal context of $T$=4 steps. Consistent with the intuition underlying TRM~\cite{trm}, this compact horizon already yields meaningful performance gains. As driving scenes evolve slowly, adjacent frames share strong visual similarity, allowing the \gls{LLM} to exploit its recurrent context more effectively rather than inferring actions from scratch at each step. While this design preserves computational efficiency and proves sufficient for the evaluated scenarios, it may be inadequate for manoeuvres requiring longer-horizon temporal reasoning. Equipping \glspl{VLA} with scalable long-horizon memory thus remains an important open challenge.

\textbf{LiDAR- and Radar-free perception.} FIVE-VLA relies exclusively on RGB camera 
input, foregoing the depth and range accuracy provided by LiDAR and Radar. While camera-only models have shown competitive performance on closed-loop benchmarks (\cref{tab:sota}), LiDAR and Radar typically add extra reliability for 3D perception in safety-critical real-world scenarios. Integrating these sensors within the VLA framework without sacrificing the efficiency gains of FIVE-VLA is an open challenge.

\blockcomment{
\renewcommand{\thesection}{A}
\input{Sections/Appendices/A0_ModelDetails}
\renewcommand{\thesection}{B}
\input{Sections/Appendices/A1_FurtherImplementationDetails}
\renewcommand{\thesection}{C}

}

\newpage

\end{document}